\documentclass[11pt,a4paper]{article}
\usepackage[utf8]{inputenc}
\usepackage{amsmath,amssymb,amsfonts,amsthm}
\usepackage{graphicx}
\usepackage{algorithm}
\usepackage{algorithmic}
\usepackage{booktabs}
\usepackage{multirow}
\usepackage{xcolor}
\usepackage{tikz}
\usepackage{pgfplots}
\usepackage{tcolorbox}
\usepackage{enumitem}
\usepackage{hyperref}

\usepackage{subcaption}
\usepackage{float}
\usepackage{dsfont}
\usepackage{fix-cm}

\usepackage[round, authoryear]{natbib}

\usepackage{arxiv}

\pgfplotsset{compat=1.17}
\usetikzlibrary{patterns,shapes,arrows}

\title{\textbf{Fortytwo: Swarm Inference with Peer-Ranked Consensus}}

\author{
 Vladyslav Larin \\
  \texttt{vlarin@fortytwo.network} \\
  \And
  Ihor Naumenko \\
  \texttt{inaumenko@fortytwo.network}
  \And
  Aleksei Ivashov \\
  \texttt{aivashov@fortytwo.network}
  \And
  Ivan Nikitin \\
  \texttt{inikitin@fortytwo.network}
  \And
  Alexander Firsov \\
  \texttt{afirsov@fortytwo.network} \\
}

\begin{document}

\maketitle

\begin{abstract}
As centralized AI hits compute ceilings and diminishing returns from ever-larger training runs, meeting demand requires an inference layer that scales horizontally in both capacity and capability. We present Fortytwo, a novel protocol that leverages swarm intelligence principles and distributed pairwise ranking consensus to achieve superior performance in AI inference. Our approach reimagines collaboration among AI nodes using \emph{swarm inference}: a peer-ranked, reputation-weighted consensus across heterogeneous models that surfaces the highest-quality responses. Using pairwise ranking with a custom Bradley--Terry–style aggregation model, we demonstrate that swarm inference substantially outperforms majority voting, achieving \textbf{85.90\%} on GPQA Diamond versus \textbf{68.69\%} for majority voting with the same model set—an improvement of \textbf{+17.21} percentage points (approximately \textbf{+25.1\%} relative). The protocol incorporates on-chain reputation so node influence adapts to demonstrated accuracy over time, yielding a meritocratic consensus that filters low-quality or malicious participants. To resist Sybil attacks, Fortytwo employs \emph{proof-of-capability} in its consensus: nodes must successfully complete calibration/test requests and \emph{stake reputation} to enter ranking rounds, making multi-identity attacks economically unattractive while preserving openness. Across six challenging benchmarks, including GPQA Diamond, LiveCodeBench, and AIME, our evaluation indicates higher accuracy and strong resilience to adversarial and noisy free-form prompting (e.g., prompt-injection degradation of only \textbf{0.12\%} versus \textbf{6.20\%} for a monolithic single-model baseline), while retaining practical deployability. Together, these results establish a foundation for decentralized AI systems—democratizing access to high-quality inference through collective intelligence without sacrificing reliability or security.
\end{abstract}

\vspace{0.5cm}

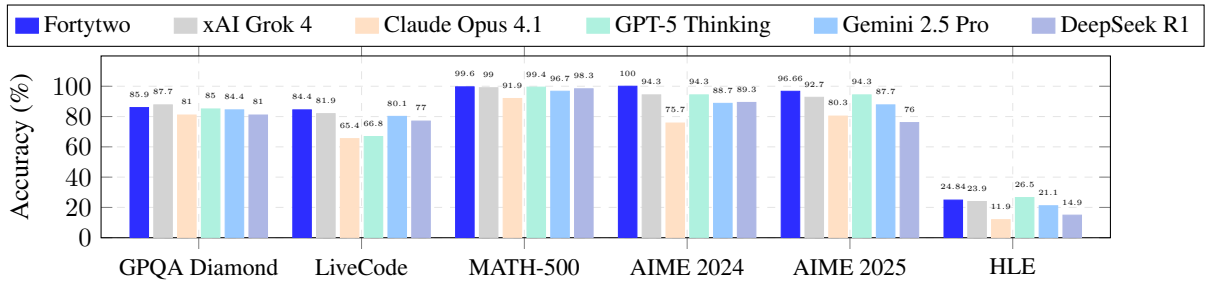
\begin{figure}[htbp]

\begin{center}
\begin{tikzpicture}

\definecolor{fortytwo}{RGB}{45, 45, 255}
\definecolor{grok}{RGB}{211, 211, 211}
\definecolor{opus}{RGB}{255, 224, 197}
\definecolor{gpt5}{RGB}{175, 241, 223}
\definecolor{gemini}{RGB}{153, 202, 255}
\definecolor{deepseek}{RGB}{174, 183, 229}

\begin{axis}[
    ybar,
    bar width=7pt,
    width=15cm,
    height=4cm,
    ylabel={Accuracy (\%)},
    ylabel style={font=\normalsize},
    symbolic x coords={GPQA Diamond, LiveCode, MATH-500, AIME 2024, AIME 2025, HLE},
    xtick=data,
    xticklabel style={font=\small},
    ymin=0, ymax=120,
    ytick={0,20,40,60,80,100},
    legend style={
        at={(0.5,1.05)},
        anchor=south,
        legend columns=6,
        font=\small,
        /tikz/every even column/.append style={column sep=0.5cm},
    },
    legend image code/.code={
            \fill[#1] (0cm,-0.1cm) rectangle (0.3cm,0.1cm);
    },
    nodes near coords,
    nodes near coords style={font=\fontsize{3}{3}\selectfont, anchor=south, yshift=0.5pt},
    enlarge x limits=0.12,
    grid=major,
    grid style={dashed,gray!20},
    ymajorgrids=true,
]

\addplot[fill=fortytwo, draw=fortytwo] coordinates {
    (GPQA Diamond, 85.90)
    (LiveCode, 84.40)
    (MATH-500, 99.60)
    (HLE, 24.84)
    (AIME 2024, 100.0)
    (AIME 2025, 96.66)
};
\addlegendentry{Fortytwo}

\addplot[fill=grok, draw=grok] coordinates {
    (GPQA Diamond, 87.70)
    (LiveCode, 81.90)
    (MATH-500, 99.00)
    (HLE, 23.90)
    (AIME 2024, 94.30)
    (AIME 2025, 92.70)
};
\addlegendentry{xAI Grok 4}

\addplot[fill=opus, draw=opus] coordinates {
    (GPQA Diamond, 81.00)
    (LiveCode, 65.40)
    (MATH-500, 91.90)
    (HLE, 11.90)
    (AIME 2024, 75.70)
    (AIME 2025, 80.30)
};
\addlegendentry{Claude Opus 4.1}

\addplot[fill=gpt5, draw=gpt5] coordinates {
    (GPQA Diamond, 85.00)
    (LiveCode, 66.80)
    (MATH-500, 99.40)
    (HLE, 26.50)
    (AIME 2024, 94.30)
    (AIME 2025, 94.30)
};
\addlegendentry{GPT-5 Thinking}

\addplot[fill=gemini, draw=gemini] coordinates {
    (GPQA Diamond, 84.40)
    (LiveCode, 80.10)
    (MATH-500, 96.70)
    (HLE, 21.10)
    (AIME 2024, 88.70)
    (AIME 2025, 87.70)
};
\addlegendentry{Gemini 2.5 Pro}

\addplot[fill=deepseek, draw=deepseek] coordinates {
    (GPQA Diamond, 81.00)
    (LiveCode, 77.00)
    (MATH-500, 98.30)
    (HLE, 14.90)
    (AIME 2024, 89.30)
    (AIME 2025, 76.0)
};
\addlegendentry{DeepSeek R1}

\end{axis}
\end{tikzpicture}
\end{center}

\caption{Benchmark performance comparison across models.}
\label{fig:benchmark_performance_short}

\end{figure}

\vspace{0.5cm}

\newpage
\tableofcontents
\newpage

\section{Introduction}\label{sec:introduction}

The landscape of artificial intelligence has undergone a fundamental transformation with the emergence of large language models (LLMs) exceeding hundreds of billions of parameters, creating unprecedented challenges in deployment, accessibility, and control \citep{agrawal2024sarathi, mafrur2025decentralized}. Traditional centralized AI infrastructures, dominated by a handful of major technology companies, have created computational bottlenecks that limit innovation and raise concerns about algorithmic transparency, data sovereignty, and equitable access to transformative AI technologies. These centralized systems face inherent limitations in scalability, with distributed inference achieving only 50-70\% of single-machine throughput due to communication overhead in wide-area deployments \citep{kwon2023efficient}. The concentration of AI capabilities within oligopolistic structures has profound implications for technological sovereignty, economic equity, and the democratic development of artificial intelligence systems that increasingly shape human society. Building on our earlier work introducing self-supervised swarm inference in trustless environments \citep{fortytwo2024}, this paper develops a complete protocol and system architecture for decentralized inference—detailing dual-role node design (generation + judging), a distributed pairwise-ranking consensus based on a custom Bradley–Terry–style aggregation model, reputation-weighted decision making, and the efficiency/robustness characteristics that emerge at swarm scale.

The emergence of decentralized AI represents a paradigm shift that addresses these fundamental limitations by distributing computation across heterogeneous networks of autonomous nodes. This approach draws inspiration from natural systems where collective intelligence emerges from simple local interactions—a phenomenon observed across biological systems from ant colonies optimizing foraging paths to neural networks processing information through distributed computation \citep{garnier2007biological, nitti2025collective}. Decentralized systems offer inherent advantages including enhanced resilience through redundancy, efficient resource utilization by leveraging idle computational capacity, and democratic access that enables participation from diverse stakeholders regardless of their economic resources. The integration of blockchain technology provides the cryptographic guarantees and economic incentives necessary for trustless coordination, creating immutable audit trails essential for regulatory compliance while enabling transparent AI decision-making \citep{castro1999practical, zhang2024byzantine}.

The technical challenges of implementing trustless AI inference have motivated numerous approaches, each representing different trade-offs in the design space of security, performance, and economic viability. Zero-Knowledge Machine Learning (ZKML) provides the strongest cryptographic guarantees by enabling verifiable inference without revealing model weights or input data, although computational overhead remains prohibitive for large-scale deployment \citep{feng2021zen, sun2023zkdl}. Optimistic Machine Learning (OPML) reduces computational requirements through fraud-proof protocols that separate execution from verification, achieving practical deployment at the cost of requiring challenge periods that delay finality \citep{conway2024opml}. Proof of Quality (PoQ) mechanisms focus on output validation using lightweight discriminator models, trading guaranteed accuracy for computational efficiency with empirical results showing less than 70\% accuracy in quality assessment \citep{zhang2024proof}. These existing approaches, while valuable contributions to the field, do not achieve the combination of accuracy, efficiency, and economic viability necessary for widespread adoption.

The theoretical foundations for our approach draw from multiple disciplines, including tournament theory, social choice theory, and swarm intelligence. The Bradley-Terry model, developed in 1952 to analyze paired comparisons in sports tournaments, provides a probabilistic framework for aggregating preferences that has found renewed relevance in AI alignment and preference modeling \citep{bradley1952rank}. Recent advances by Sun et al. provide statistical consistency and convergence guarantees for Bradley–Terry-style preference models in neural settings, giving formal support to our use of pairwise ranking as a foundation for decentralized evaluation \citep{sun2024rethinking}. The model's ability to extract consistent global rankings from potentially inconsistent pairwise comparisons makes it particularly suitable for distributed evaluation, where individual nodes may have varying reliability and expertise.

The principles of swarm intelligence, observed in biological systems and increasingly applied to artificial systems, demonstrate how collective behaviors can emerge from simple local rules without centralized coordination \citep{dorigo2021swarm}. Natural swarms achieve remarkable feats of collective problem solving: ant colonies discover optimal paths through pheromone-based stigmergic communication, bee colonies make democratic decisions about nest sites through waggle dance debates, and bird flocks navigate complex environments through local alignment rules \citep{garnier2007biological}. These biological systems have evolved mechanisms for robust decision making under uncertainty, resistance to individual failures, and adaptation to changing environments, essential properties for artificial intelligence systems operating in real-world conditions.

\subsection{Multidisciplinary Foundations}\label{sec:multidisciplinary}

To ensure accessibility for researchers from diverse backgrounds, we provide comprehensive explanations of key concepts that underpin our framework. The Bradley-Terry model operates on the principle that the probability of preferring the item $i$ over the item $j$ follows a logistic function based on latent quality scores: $P(i \succ j) = \pi_i/(\pi_i + \pi_j)$, where $\pi_i$ represents the intrinsic quality of the item $i$. This formulation naturally handles intransitive preferences and provides uncertainty estimates through the variance of score distributions. In the context of AI systems, each response generated by a participant model can be assigned a quality score based on pairwise comparisons with other responses, allowing robust ranking even when direct quality measurement is challenging.

Byzantine Fault Tolerance (BFT), a fundamental concept in distributed systems, addresses the challenge of achieving consensus when some participants may behave arbitrarily or maliciously, named after the Byzantine Generals Problem, where commanders must coordinate despite potential traitors \citep{lamport1982byzantine}. Classical results establish that BFT protocols can tolerate up to $f < n/3$ Byzantine nodes among $n$ total nodes, with modern implementations such as HotStuff achieving linear message complexity suitable for internet-scale deployment \citep{yin2019hotstuff}. Our system leverages these theoretical guarantees while introducing reputation-based weighting that provides additional resilience beyond classical fault tolerance bounds.

The reputation system in our framework serves as a dynamic weighting mechanism where each nodes's influence on consensus decisions evolves based on their historical performance. Unlike static voting systems where all participants have equal weight, reputation-based systems create meritocracies where consistent quality contributions are rewarded with increased influence. The reputation $R_i$ of node $i$ is updated according to: $R_{t+1}^{(i)} = \alpha R_t^{(i)} + (1-\alpha) \cdot \text{Accuracy}_t^{(i)}$, where $\alpha$ controls the balance between the historical reputation and recent performance. This creates powerful incentives for honest participation, while naturally marginalizing low-quality or malicious nodes over time.

\subsection{Key Innovations and Contributions}

This paper presents the Fortytwo Protocol, a comprehensive framework that synthesizes insights from swarm intelligence, tournament theory, and distributed systems to create a practical solution for decentralized AI inference. Our key innovations include:

\begin{enumerate}
    \item \textbf{Distributed Pairwise Ranking Consensus:} We introduce a novel mechanism where each node generates $3N$ pseudo-random pairwise comparisons that are aggregated using \emph{our improved Bradley--Terry model}. This leverages the cognitive principle that relative comparisons are more reliable than absolute assessments, while our Bradley--Terry extension ensures convergence to consistent global rankings even under noisy or inconsistent individual comparisons.
    
    \item \textbf{Multi-Token Reasoning Chains:} Unlike single-scalar reward models that compress quality assessment into a single forward pass, our system requires nodes to generate detailed reasoning chains (50-100 tokens) explaining their ranking decisions. This explicit reasoning process improves accuracy by 5.3\% compared to single-token scoring while creating audit trails that allow system improvement and debugging.
    
    \item \textbf{Compute Stake Mechanism:} We introduce a compute-anchored Sybil-resistance scheme that relies on \emph{proof of capability} rather than economic staking. New nodes must complete comprehensive test calls across their claimed semantic distribution (e.g., mathematical reasoning, scientific analysis, code generation) to demonstrate competence, with the computational cost of these evaluations serving as a natural barrier to identity multiplication.
    
    \item \textbf{Adaptive Reputation Dynamics:} The system implements sophisticated reputation tracking where node's weights evolve based on ranking accuracy, consistency, and alignment with consensus. Poor performers experience reputation "slashing" analogous to stake slashing in proof-of-stake systems, while consistent quality contributions are rewarded with increased influence and higher reward shares.
    
    \item \textbf{Adversarial Resilience and Free-form Stability:} Swarm Inference directly mitigates \emph{contextual distraction}—the tendency of single LLMs to be derailed by extraneous or misleading context. By combining ensemble diversity with peer-ranked validation, the network preserves signal under noisy, unstructured, or intentionally manipulative prompts (e.g., prompt injections, verbose or poorly formatted inputs, mixed domains). In our evaluations, accuracy degraded by only \textbf{0.12\%} under such noisy/adversarial prompting, versus \textbf{6.20\%} for a monolithic single-model baseline. This robustness translates to higher real-world accuracy on free-form user queries and exhibits antifragile behavior: diversity and consensus reduce failure modes as the swarm grows.
\end{enumerate}

\subsection{Paper Organization}

The remainder of this paper is organized as follows: Section 2 provides a comprehensive survey of related work in decentralized AI, establishing the context and limitations of existing approaches. Section 3 presents the detailed architecture of self-supervised inference, including the dual-role design where nodes perform both inference and ranking. Section 4 describes our enhanced pairwise ranking consensus mechanism, including the foundational Bradley–Terry aggregation framework, and reputation-weighted voting. Section 5 details the compute stake mechanism and its effectiveness in preventing Sybil attacks. Section 6 presents extensive experimental evaluation across multiple benchmarks, demonstrating superior performance and adversarial robustness. Section 7 analyzes why our approach succeeds, exploring the cognitive, statistical, and economic factors that contribute to its effectiveness. Section 8 discusses limitations and future research directions. Finally, Section 9 concludes with reflections on the broader implications for decentralized AI development.

\section{Related Work}

The pursuit of trustless AI inference in decentralized environments has spawned diverse approaches, each addressing different aspects of the fundamental trilemma between security, performance, and economic viability. We systematically examine these approaches, analyzing their contributions and limitations to establish the context for our work.

\subsection{Cryptographic Approaches to Verifiable AI}

\subsubsection{Zero-Knowledge Machine Learning}

Zero-Knowledge Machine Learning represents the standard for cryptographic verifiability, enabling proof of correct inference without revealing model weights or input data. The foundational work by Feng et al. introduced Zen, an optimizing compiler that transforms neural network computations into arithmetic circuits suitable for zero-knowledge proof systems \citep{feng2021zen}. This approach achieves perfect security guarantees—a verifier can be certain that the claimed inference was computed correctly without learning anything about the model or data beyond the output.

Recent advances have focused on improving practical efficiency. The EZKL framework provides production-ready tools for converting ONNX models to zkSNARK circuits using the Halo2 proof system, supporting networks up to 100 million parameters \citep{ezkl2024framework}. Daniel Kang et al. demonstrated scaling to larger models through specialized optimizations, though proof generation for a single ResNet-50 inference still requires over 24 hours \citep{kang2022scaling}. The fundamental challenge remains the superlinear growth of proof generation complexity with model size—each multiplication in the neural network requires multiple constraints in the arithmetic circuit, and proof generation time grows polynomially with circuit size.

The computational overhead of ZKML makes it suitable only for high-value applications where cryptographic guarantees justify extreme costs. Financial fraud detection, medical diagnosis verification, and legal evidence validation represent domains where the ability to prove computational correctness without revealing proprietary models or sensitive data provides sufficient value to offset computational expenses. However, for general-purpose AI inference where users expect interactive response times, ZKML remains impractical despite its theoretical elegance.

\subsubsection{Homomorphic Encryption}

Fully Homomorphic Encryption (FHE) enables computation on encrypted data, providing strong privacy guarantees for distributed inference. Stoian et al. constructed deep neural network architectures compatible with TFHE (Torus Fully Homomorphic Encryption), achieving practical inference on encrypted MNIST data with accuracy comparable to plaintext models \citep{stoian2023deep}. The approach ensures that model providers never see user data while users never access model weights, addressing bidirectional privacy concerns in AI-as-a-service deployments.

Recent advances in CKKS-based schemes support approximate arithmetic operations essential for neural network computations, with improvements in ciphertext packing and bootstrapping that reduce amortized costs \citep{hong2025recent}. Microsoft's SEAL library and IBM's HELayers provide production-ready implementations, though performance remains challenging: a single encrypted inference on a miniature ResNet-20 requires seconds to minutes compared to microseconds for plaintext computation. The multiplicative depth limitations of practical FHE schemes also constrain model architectures, preventing the use of deep networks that have driven recent AI advances.

\subsection{Optimistic and Economic Security Models}

\subsubsection{Optimistic Machine Learning}

Optimistic Machine Learning represents a pragmatic compromise between security and efficiency, adopting the "optimistic" assumption that computations are correct unless challenged. Conway et al. introduced a comprehensive OPML architecture featuring Fraud Proof Virtual Machines and Interactive Dispute Games that enable practical deployment of 7B-parameter models \citep{conway2024opml}. The system separates execution from proving—inference results are available immediately while cryptographic proofs are generated only when disputes arise.

The economic security model creates powerful disincentives for misbehavior: challengers must stake tokens that are forfeited if disputes prove frivolous, while service providers lose stakes if caught cheating. This mechanism has proven effective in practice, with ORA Protocol's on-chain AI Oracle processing millions of inferences without successful attacks. The key limitation is the challenge period (typically 7 days) required for finality, making OPML unsuitable for applications requiring immediate quality guarantee, that is, majority of practical applications. Additionally, the complexity of implementing fraud proofs for neural network computations requires sophisticated engineering to handle numerical precision issues and non-deterministic operations.

\subsubsection{Proof of Quality Mechanisms}

Proof of Quality approaches shift focus from computational correctness to output quality, using lightweight discriminator models to evaluate generated content. Zhang et al. developed comprehensive frameworks where simple BERT-based models assess outputs from complex generative models, achieving validation with less than 1\% of generation computational cost \citep{zhang2024proof}. This asymmetry between generation and validation costs makes PoQ economically attractive for decentralized systems.

Chong et al. extended this concept with Proof of Useful Intelligence, demonstrating that quality-based consensus can achieve 97\% energy reduction compared to Proof-of-Work while maintaining security through economic incentives \citep{chong2025proof}. However, fundamental challenges remain in quality assessment accuracy—empirical evaluations show discriminator models achieve less than 70\% agreement with human judgments for complex outputs. The subjective nature of quality in creative tasks further complicates validation, as different evaluators may have legitimate disagreements about output quality.

\subsection{Distributed Consensus and Byzantine Fault Tolerance}

\subsubsection{Classical Byzantine Fault Tolerance}

Byzantine Fault Tolerance provides the theoretical foundation for consensus in adversarial environments. The Byzantine Generals Problem, formulated by Lamport, Shostak, and Pease, establishes fundamental limits: any deterministic protocol requires $n \geq 3f + 1$ nodes to tolerate $f$ Byzantine failures \citep{lamport1982byzantine}. This bound is tight—no protocol can tolerate more failures, and protocols exist that achieve consensus with exactly this many nodes.

Practical Byzantine Fault Tolerance (PBFT), introduced by Castro and Liskov, made BFT practical for real systems by reducing message complexity to $O(n^2)$ and achieving throughput of thousands of transactions per second \citep{castro1999practical}. The protocol operates in rounds where a designated primary node proposes values that backup nodes validate through three-phase voting. Despite its efficiency, PBFT suffers from poor scalability beyond hundreds of nodes due to quadratic message complexity and the overhead of view changes when primaries fail.

\subsubsection{Modern BFT Innovations}

Recent advances address classical BFT limitations through various approaches. HotStuff achieves linear message complexity $O(n)$ by introducing a novel three-phase voting structure with pipelined rounds, enabling responsive operation where the protocol proceeds at network speed rather than waiting for timeout \citep{yin2019hotstuff}. This improvement makes BFT practical for internet-scale deployments with thousands of participants.

Probabilistic approaches trade deterministic guarantees for improved efficiency. Avalanche introduces metastable consensus through repeated random sampling, achieving agreement with high probability while requiring only logarithmic message complexity \citep{teamrocket2019avalanche}. These protocols are particularly suitable for AI applications where occasional inconsistencies are acceptable if they occur with negligible probability.

\subsection{Ranking and Preference Learning Systems}

\subsubsection{Bradley-Terry Model Applications}

The Bradley-Terry model has emerged as fundamental for preference-based learning in AI systems. Originally developed for sports rankings, the model assumes pairwise preference probabilities follow a logistic function based on latent quality scores \citep{bradley1952rank}. Sun et al. derive finite-sample guarantees for maximum-likelihood estimation under Bradley–Terry with neural parameterizations, specifying identifiability and comparison-graph connectivity conditions under which true latent scores are recovered with high probability, along with convergence rates \citep{sun2024rethinking}.

Applications in AI include ChatGPT's reward modeling where human preferences between response pairs train reward functions for reinforcement learning \citep{ouyang2022training}. The Chatbot Arena uses Bradley-Terry modeling to rank language models based on crowd-sourced pairwise comparisons, producing leaderboards that correlate strongly with comprehensive benchmarks \citep{chiang2024chatbot}. These successes demonstrate the model's effectiveness in aggregating subjective preferences into consistent global rankings.

\subsubsection{LLM-as-a-Judge Paradigms}

Recent advances demonstrate that language models can effectively evaluate their own and others' outputs when provided with appropriate prompting strategies. MetaRanking approaches demonstrate that even smaller models like Phi-2 (2.7B parameters) can provide effective ranking when fine-tuned on preference data \citep{liu2024meta}. The ability to perform accurate ranking with smaller models enables broader participation in consensus formation, as nodes with limited computational resources can still contribute meaningful evaluations. Zheng et al. established MT-Bench as a widely-used benchmark for LLM evaluation, achieving 0.93 Spearman correlation with human annotators and validating the reliability of automated assessment \citep{zheng2023judging}.

\subsection{Swarm Intelligence and Collective Systems}

\subsubsection{Biological Inspiration}

Swarm intelligence emerges from simple local interactions among nodes following basic rules without centralized coordination. Dorigo's foundational work on Ant Colony Optimization demonstrated how artificial ants could solve complex optimization problems through pheromone-based communication \citep{dorigo1996ant}. These algorithms have found practical applications in network routing, scheduling, and resource allocation where traditional optimization methods struggle with dynamic environments.

Foundational research by Garnier et al. establishes the biological principles underlying swarm coordination, identifying key mechanisms including positive feedback for rapid decision-making, negative feedback for stabilization, randomness for exploration, and multiple interactions for information averaging \citep{garnier2007biological}. These principles directly inform our node coordination mechanisms, particularly in how the ranking information propagates through the swarm to achieve consensus as well how nodes use random sampling parameters or aggregate on the best candidates.

\subsubsection{Artificial Swarm Intelligence}

The application of swarm principles to human and AI groups has demonstrated remarkable potential for amplifying collective intelligence. Rosenberg’s Artificial Swarm Intelligence (ASI) platform connects groups of human participants in real time, allowing them to form dynamic feedback loops modeled after biological swarms. In one study, 75 sports fans working together as a real-time swarm on the UNU platform achieved 70\% accuracy in predicting College Bowl football games against the spread—outperforming both individual participants (50\%) and expert analysts (50\%) \citep{rosenberg2016artificial}. By synchronizing individual inputs through continuous feedback and convergence mechanisms, the system enables human groups to reach decisions that consistently exceed the predictive power of traditional polling and voting methods.

Recent implementations in robotics demonstrate practical swarm coordination. Nitti et al. developed swarm cooperation models achieving higher or equal success rate with respect to benchmark methods on 22 out of 33 landscapes with groups of 5-16 robots through multi-loop feedback and adaptive control \citep{nitti2025collective}. These results suggest that modest-sized swarms can achieve most benefits of larger collectives, informing our design decision to focus on swarms of 5-19 nodes that balance performance with coordination overhead.

\subsection{Security and Attack Prevention}

\subsubsection{Sybil Attack Prevention}

Sybil attacks, where adversaries create multiple fake identities to gain disproportionate influence, represent a fundamental threat to decentralized systems \citep{douceur2002sybil}. Traditional defenses rely on resource costs: Proof-of-Work requires computational expenditure making identity creation expensive, while Proof-of-Stake demands economic investment proportional to influence. However, these mechanisms favor participants with greater capital resources, conflicting with goals of democratization and efficiency.

Recent approaches explore alternative Sybil defenses. Social network-based methods like SybilGuard leverage trust relationships to limit fake identities, though assumptions about social graph properties may not hold in practice \citep{yu2008sybilguard}. Proof of Personhood systems attempt to ensure one-person-one-vote through mechanisms like pseudonym parties or biometric verification, though privacy and usability challenges remain \citep{ford2017proof}. Our compute stake mechanism represents a novel approach requiring demonstration of capability rather than resource expenditure.

\subsubsection{Adversarial Robustness}

Adversarial attacks on AI systems have evolved from simple perturbations fooling image classifiers to advanced prompt injections manipulating language models. Recent work demonstrates that ensemble methods provide natural defense—attacks that fool one model often fail against others with different architectures or training \citep{pang2021improving}. This diversity principle fundamentally underlies our swarm's resilience to adversarial inputs.

Research on certified robustness provides theoretical guarantees about model behavior under bounded perturbations, though current methods apply only to simple models and small perturbation bounds \citep{wong2018provable}. Our approach achieves practical robustness through redundancy and validation rather than certification, demonstrating only 0.12\% performance degradation under adversarial conditions compared to 6.20\% for individual models.

\section{Self-Supervised Inference Architecture}

The design of our self-supervised inference represents a fundamental departure from traditional distributed AI architectures where separate entities handle inference and validation. By empowering each node with dual capabilities for both content generation and quality assessment, we create a self-organizing ecosystem where collective intelligence emerges from local interactions without centralized coordination. This section presents the comprehensive architecture, theoretical foundations, and implementation details of our node design.

\subsection{Theoretical Foundations of Dual-Role Nodes}

The decision to combine inference and ranking capabilities within single nodes stems from multiple theoretical and practical considerations. From a game-theoretic perspective, nodes that both generate content and evaluate others' outputs have "skin in the game"—their ranking performance directly affects their reputation and future earnings, creating powerful incentives for honest evaluation. This alignment of interests contrasts with systems where separate validators have no stake in generation quality, potentially leading to lazy or adversarial validation.

The phenomenon of emergent metacognition in large language models provides theoretical support. Recent studies show that LLMs develop implicit models of their own capabilities and limitations, enabling them to assess uncertainty and identify potential errors in reasoning \citep{kadavath2022metacognition}. This metacognitive ability naturally extends to evaluating others' outputs, as the same quality criteria apply regardless of the source. By leveraging this inherent capability, our dual-role design achieves efficiency and consistency that would be difficult to replicate with separate generation and validation systems.

\subsection{Node Component Architecture}

Each node in our swarm implements a modular architecture that balances flexibility with standardization. The design enables heterogeneous implementations while maintaining compatible interfaces for inter-node communication and system integration.

\subsubsection{Primary Cognitive Module}

The core component responsible for both content generation and ranking tasks can be implemented using various approaches depending on available resources and specialization requirements:

\paragraph{Large Language Models:} The most common implementation uses transformer-based language models that typically range from 3B to 405B parameters. Smaller models like Qwen3-8B or Gemma-3-4B provide efficient operation on consumer hardware while maintaining competitive performance. Larger models like gpt-oss-120b, Qwen3-235B or DeepSeek R1 offer superior capabilities, but require enterprise-grade infrastructure. The choice of model size represents a trade-off between quality and resource requirements, with our experiments showing that diverse swarms mixing model sizes often outperform homogeneous deployments of either small or large models alone.

\paragraph{Specialized Expert Systems:} Domain-specific applications benefit from specialized models fine-tuned or designed for particular tasks. Mathematical reasoning nodes might employ models trained on theorem proving and symbolic manipulation. Code generation nodes could use models fine-tuned on programming languages and software engineering practices. Scientific analysis nodes might integrate domain-specific knowledge bases and reasoning capabilities. These specialized nodes contribute unique expertise to the swarm's collective intelligence.

\paragraph{Hybrid Architectures:} Some nodes combine neural and symbolic approaches to leverage the strengths of both paradigms. Neural components handle natural language understanding and generation while symbolic reasoners ensure logical consistency and perform precise computations. This hybrid approach is particularly valuable for tasks requiring both creativity and accuracy, such as mathematical problem-solving or scientific hypothesis generation.

\paragraph{Ensemble Models:} Individual nodes can internally implement ensemble methods, combining predictions from multiple smaller models to achieve robustness and accuracy exceeding any single component. This nested ensemble structure—ensembles within a swarm—provides multiple layers of diversity and error correction. Techniques like mixture-of-experts architectures enable efficient ensemble operation by activating only relevant experts for each input.

\subsubsection{Auxiliary Processing Unit}

The optional auxiliary processing unit augments primary cognitive capabilities through specialized operations that enhance response quality and enable integration with external resources:

\paragraph{Pre-processing Pipeline:} Input text undergoes pre-processing including tokenization, prompt engineering to optimize model performance, context window management to handle long inputs efficiently, and validation to detect and filter malicious inputs. Advanced techniques like retrieval-augmented generation (RAG) and/or WebSearch can be implemented here, enriching inputs with relevant information from external knowledge bases.

\paragraph{Post-processing Operations:} Generated outputs benefit from various enhancements including grammar and style correction for improved readability, fact-checking against authoritative sources when applicable, consistency validation to ensure logical coherence, code analyzers and checkers for better interpretability of code artifacts, and output sanitization to remove potentially harmful content. These operations can significantly improve response quality without requiring larger models, making them particularly valuable for resource-constrained nodes.

\paragraph{External Tool Integration:} Nodes can leverage external tools and services through standardized interfaces. Mathematical nodes might use computer algebra systems for symbolic computation. Programming nodes could execute code in sandboxed environments for validation. Research nodes might query scientific databases and literature repositories. This tool use capability enables nodes to overcome inherent LLM limitations in specific domains while maintaining the flexibility of natural language interfaces.

\paragraph{Caching and Optimization:} Caching mechanisms reduce redundant computations by storing and reusing results for similar queries. Semantic similarity-based cache lookup enables approximate matching that captures conceptual equivalence beyond exact string matching. Dynamic cache management balances memory usage with hit rates.

\subsubsection{Ranking Engine}

The ranking engine implements our novel pairwise comparison algorithm with several features ensuring robust and manipulation-resistant evaluation:

\paragraph{Comparison Generation:} Each node generates up to $3N$ random pairwise comparisons from the response set, excluding its own submissions to prevent self-promotion bias. The randomization uses cryptographically secure random number generation seeded with blockchain state to ensure unpredictability while maintaining reproducibility for audit purposes. The number $3N$ is carefully chosen based on empirical analysis that shows that it's sufficient for robust Bradley-Terry estimation while avoiding excessive computational overhead.

\paragraph{Multi-Token Reasoning:} For each comparison, nodes generate detailed explanations (50-100 tokens) articulating their ranking rationale. This requirement serves multiple purposes: it forces systematic evaluation reducing arbitrary decisions, creates audit trails enabling system improvement, and provides training data for improving future ranking models. The reasoning must address specific quality dimensions including technical accuracy, completeness, coherence, and relevance, ensuring comprehensive evaluation rather than single-factor decisions.

\paragraph{Bradley--Terry Optimization:} The ranking engine builds upon an extended Bradley--Terry score estimation method using gradient ascent with adaptive learning rates. Regularization mitigates overfitting to noisy comparisons while preserving sensitivity to genuine quality differences. The optimization process incorporates convergence detection to eliminate redundant computation and employs numerical stabilization techniques to handle extreme score disparities.

\paragraph{Adversarial Robustness:} Multiple mechanisms protect against ranking manipulation including detection of suspicious patterns (e.g., always ranking certain nodes highly), statistical analysis to identify outlier rankings, reputation penalties for consistently poor ranking performance, and cross-validation where nodes' rankings are compared against held-out ground truth. These defenses make it economically unviable to attempt systematic manipulation while allowing for legitimate disagreements and diverse perspectives.

\subsubsection{Communication Module}

The communication module handles all inter-node messaging and coordination, implementing protocol for distributed consensus formation:

\paragraph{Gossip Protocol:} Nodes use epidemic-style gossip protocols for efficient information dissemination without requiring global broadcast. Each node periodically exchanges state with randomly selected peers, ensuring information eventually reaches all participants with high probability. The protocol includes mechanisms for detecting and reconciling conflicting information, managing network partitions and rejoining, and prioritizing critical messages during high load. Gossip protocols provide excellent scalability and fault tolerance, crucial for maintaining system operation despite individual node failures.

\paragraph{Encrypted Messaging:} All inter-node communication uses end-to-end encryption to preserve privacy and prevent tampering. The encryption scheme supports both point-to-point secure channels for sensitive exchanges and broadcast encryption for efficient group communication. Threshold cryptography enables messages readable only when sufficient nodes collaborate, useful for revealing responses only after all submissions are complete. Perfect forward secrecy ensures that compromise of current keys doesn't expose past communications.

\paragraph{Blockchain Coordination:} Critical consensus events are recorded on-chain for transparency and immutability. Smart contracts coordinate the consensus protocol, ensuring all nodes follow the same rules and timeline. The blockchain serves as a source of randomness for comparison generation, a timestamping service for ordering events, and a dispute resolution mechanism when nodes disagree.

\paragraph{Fault Tolerance:} The communication module implements comprehensive fault tolerance mechanisms including timeout-based failure detection with adaptive thresholds, automatic failover to backup communication channels, message retransmission with exponential backoff, and Byzantine fault detection through message validation. These mechanisms ensure the system continues operating despite network failures, node crashes, or deliberate attacks on communication infrastructure.

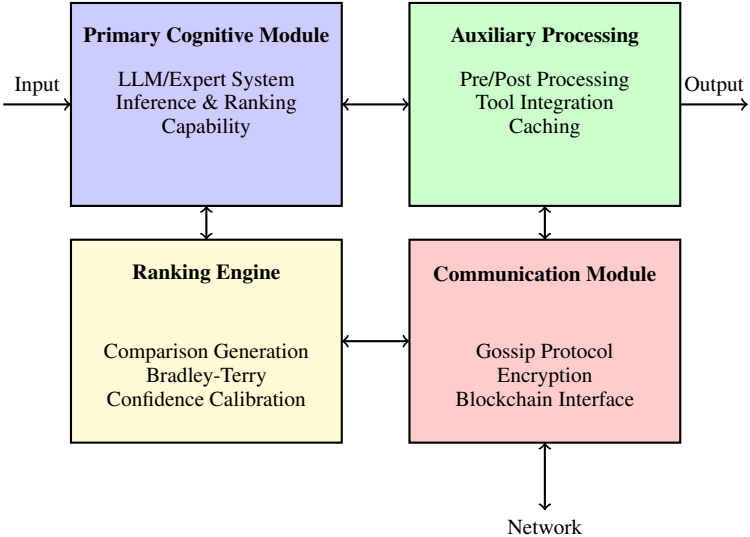
\begin{figure}[h]
\centering
\begin{tikzpicture}[scale=0.9, every node/.style={scale=0.9, font=\small}]
    \draw[thick, fill=blue!20] (0,0) rectangle (4,3);
    \node at (2,2.5) {\textbf{Primary Cognitive Module}};
    \node[align=center] at (2,1.5) {LLM/Expert System\\Inference \& Ranking\\Capability};
    
    \draw[thick, fill=green!20] (5,0) rectangle (9,3);
    \node at (7,2.5) {\textbf{Auxiliary Processing}};
    \node[align=center] at (7,1.5) {Pre/Post Processing\\Tool Integration\\Caching};
    
    \draw[thick, fill=yellow!20] (0,-3.5) rectangle (4,-0.5);
    \node at (2,-1) {\textbf{Ranking Engine}};
    \node[align=center] at (2,-2.5) {Comparison Generation\\Bradley-Terry\\Confidence Calibration};
    
    \draw[thick, fill=red!20] (5,-3.5) rectangle (9,-0.5);
    \node at (7,-1) {\textbf{Communication Module}};
    \node[align=center] at (7,-2.5) {Gossip Protocol\\Encryption\\Blockchain Interface};
    
    \draw[thick, <->] (4,1.5) -- (5,1.5);
    \draw[thick, <->] (2,0) -- (2,-0.5);
    \draw[thick, <->] (7,0) -- (7,-0.5);
    \draw[thick, <->] (4,-2) -- (5,-2);
    
    \draw[thick, ->] (-1,1.5) -- (0,1.5) node[midway, above] {Input};
    \draw[thick, ->] (9,1.5) -- (10,1.5) node[midway, above] {Output};
    \draw[thick, <->] (7,-3.5) -- (7,-4.5) node[below] {Network};
\end{tikzpicture}
\caption{Modular architecture of a self-supervised inference showing the four key components and their interactions}
\label{fig:node_architecture}
\end{figure}

\subsection{Dynamic Node Lifecycle}

The nodes in our system undergo a lifecycle that ensures quality while maintaining openness to new participants. It creates natural selection pressure that favors competent and honest nodes while filtering out poor performers.

\subsubsection{Entry and Qualification}

New nodes must demonstrate competence through comprehensive test/free participation before participating in paid consensus rounds. Nodes must achieve minimum accuracy thresholds in each claimed domain, ensuring well-rounded capability rather than narrow specialization. The computational cost of completing these tests serves as our novel compute stake mechanism, creating a natural barrier against Sybil attacks. The qualification process adapts according to the current network needs. When the system lacks expertise in specific domains, the qualification requirements are naturally relaxed for specialists in those areas. Conversely, oversupplied capabilities face higher barriers to entry. 

\subsubsection{Active Participation}

Qualified nodes participate in consensus rounds based on their reputation and availability. Higher-reputation nodes receive priority for high-value requests and earn larger reward shares. The system implements fair scheduling ensuring all qualified nodes receive opportunities to participate, preventing reputation-rich nodes from monopolizing the system. Load balancing distributes requests across nodes based on their current utilization, maintaining system responsiveness.

During active participation, nodes continuously learn and adapt. They can observe which responses win consensus and why, enabling improvement through self-reflection. Nodes might fine-tune their models on successful responses or adjust their ranking strategies based on observed patterns. This continuous learning creates evolutionary pressure toward better performance, with the swarm as a whole becoming more capable over time.

\subsubsection{Reputation Evolution}

The reputation of each node evolves through a dynamic process that rewards consistent quality and penalizes poor performance. The reputation update mechanism considers multiple factors, including the accuracy of generated responses (how often a node’s submissions win consensus), the quality of ranking decisions (correlation with final consensus rankings), consistency across different query types, and timeliness in meeting deadlines. The system employs exponential moving averages to balance stability with responsiveness to performance changes.

Reputation decay ensures that nodes must maintain active participation to preserve their standing. Inactive nodes gradually lose reputation, preventing the accumulation of influence by early participants who stop contributing. This mechanism also provides a recovery path for nodes that experience temporary performance degradation, as consistent good performance can rebuild lost reputation over time.

\subsubsection{Exit and Penalties}

Nodes whose reputation falls below minimum thresholds face computation-stake slashing — a penalty mechanism in which accumulated computational reputation and participation rights are reduced or revoked. Slashing, a concept borrowed from Proof-of-Stake consensus systems, refers to the forfeiture of staked assets as punishment for protocol violations or poor performance. In our system, the “stake” consists of the computational work invested in capability demonstrations and the reputation earned through quality contributions, rather than financial tokens.

This mechanism establishes a meaningful barrier, making the cost of circumventing exclusion prohibitive relative to the benefits of honest participation.

\subsection{Semantic Network Topology}

The physical organization of nodes in our swarm leverages a semantic embedding–based topology that enables efficient discovery and routing based on capability similarity. This approach moves beyond random or geographic organization, forming a self-organizing network in which nodes with related expertise naturally cluster together.

\subsubsection{Semantic Embedding Space}

Each node $a_i$ maintains a set of semantic embeddings $\{\mathbf{x}_{i,j}\} \subset \mathbb{R}^d$ representing its capabilities across different domains, where $d$ is the embedding dimension (typically 768 or 1536 for modern language models). These embeddings are derived from:

\begin{itemize}
    \item \textbf{Capability vectors:} Generated from the node's model architecture and fine-tuning
    \item \textbf{Historical performance:} Weighted averages of successfully answered query embeddings
    \item \textbf{Declared specializations:} Explicit domain tags specified by operator
\end{itemize}

The semantic distance between nodes is computed using cosine similarity in the embedding space:

\begin{equation}
d_{\text{sem}}(a_i, a_j) = 1 - \max_{k,l} \frac{\mathbf{x}_{i,k} \cdot \mathbf{x}_{j,l}}{\|\mathbf{x}_{i,k}\| \|\mathbf{x}_{j,l}\|}
\end{equation}

\subsubsection{Axis-Aligned Hierarchical Partitioning}

To efficiently organize nodes in the high-dimensional semantic space, we employ an axis-aligned partitioning strategy inspired by k-d trees. This approach recursively subdivides the space to create balanced sub-meshes:

\begin{algorithm}
\caption{Semantic Space Partitioning}
\begin{algorithmic}[1]
\STATE Initialize root region $\Omega \subset \mathbb{R}^d$ containing all node embeddings
\WHILE{$\exists$ region $\Omega_k$ with $|\text{nodes}(\Omega_k)| > \beta$ or $\text{load}(\Omega_k) > \lambda$}
    \STATE Select dimension $\ell^* = \arg\max_{\ell} \text{Var}_{\ell}(\Omega_k)$ with maximum variance
    \STATE Compute median $m_{\ell^*}$ along dimension $\ell^*$
    \STATE Split $\Omega_k$ into $\Omega_{k,\text{left}}$ and $\Omega_{k,\text{right}}$ at hyperplane $x_{\ell^*} = m_{\ell^*}$
    \STATE Create sub-meshes $M_{\text{left}}$ and $M_{\text{right}}$ from nodes in respective regions
\ENDWHILE
\RETURN Hierarchical partition tree with leaf sub-meshes
\end{algorithmic}
\end{algorithm}

For the nodes that lay on the median we use id-dependent pseudorandom to choose the left or right side for each ambiguous median.

This partitioning strategy offers several advantages over uniform grid-based approaches:

\begin{itemize}
    \item \textbf{Balanced load distribution:} Each split ensures roughly equal numbers of nodes in child regions
    \item \textbf{Adaptive granularity:} High-traffic regions are subdivided more finely
    \item \textbf{Dimension-aware splitting:} Choosing the axis of maximum variance prevents degenerate partitions
    \item \textbf{Scalability:} $O(\log n)$ depth ensures efficient routing even with millions of nodes
\end{itemize}

\subsubsection{Dynamic Sub-Mesh Formation}

Sub-meshes are dynamically created and adjusted based on observed request patterns. Each sub-mesh $M_k$ maintains:

\begin{equation}
M_k = \{a_i \in A : \exists j, \mathbf{x}_{i,j} \in \Omega_k\}
\end{equation}

where $\Omega_k$ is the semantic region defining the sub-mesh. The system monitors request rates $R(\Omega_k, t)$ and triggers subdivision when:

\begin{equation}
R(\Omega_k, t) = \sum_{a_i \in M_k} r_i(t) > \lambda_{\text{split}}
\end{equation}

This adaptive mechanism ensures that high-demand semantic regions receive proportionally more resources and finer-grained organization.

\subsubsection{Bandwidth-Aware Coverage Radius}

Each node $a_i$ with semantic point $\mathbf{x}_{i,j}$ maintains a coverage radius $r_{i,j}$ that defines its responsibility region in semantic space. The radius is dynamically adjusted to respect bandwidth constraints, thus each node concentrates on areas of maximum competence:

\begin{equation}
r_{i,j}^{(t+1)} = \begin{cases}
r_{i,j}^{(t)} \cdot (1 + \delta) & \text{if } \sum_{\mathbf{x} \in B(\mathbf{x}_{i,j}, r_{i,j})} r(\mathbf{x}, t) < \alpha \theta_i \\
r_{i,j}^{(t)} \cdot (1 - \delta) & \text{otherwise}
\end{cases}
\end{equation}

where $B(\mathbf{x}_{i,j}, r)$ denotes the ball of radius $r$ centered at $\mathbf{x}_{i,j}$, $\theta_i$ is node $i$'s bandwidth limit, and $\alpha < 1$ provides a safety margin.

\subsubsection{Efficient Query Routing}

When a new query $q$ arrives, the system routes it through the semantic topology:

\begin{enumerate}
    \item Compute query embedding $\mathbf{q} = \text{encode}(q)$
    \item Traverse the partition tree to find leaf region $\Omega_k$ containing $\mathbf{q}$
    \item Broadcast to all nodes from sub-mesh $M_k$
\end{enumerate}

This hierarchical routing achieves $O(\log n)$ complexity compared to $O(n)$ for exhaustive search, enabling real-time node selection even in large swarms.

\subsubsection{Sub-Mesh Identification and Coordination}

Each sub-mesh receives a unique identifier encoding its path in the partition tree:

\begin{equation}
\text{ID}(M_k) = \ell_1\text{-}d_1 \parallel \ell_2\text{-}d_2 \parallel \ldots \parallel \ell_p\text{-}d_p
\end{equation}

where $\ell_i \in \{L, R\}$ indicates left/right splits and $d_i$ specifies the splitting dimension. This hierarchical naming enables efficient routing table construction and sub-mesh discovery.

The semantic topology provides several key benefits to our swarm consensus mechanism:

\begin{itemize}
    \item \textbf{Expertise clustering:} Nodes with similar capabilities naturally group together, improving the quality of local consensus
    \item \textbf{Load balancing:} Dynamic partitioning prevents hotspots by distributing high-demand regions across multiple sub-meshes
    \item \textbf{Efficient discovery:} Logarithmic routing complexity enables rapid node selection for consensus rounds
    \item \textbf{Semantic locality:} Queries are primarily handled by semantically relevant nodes, improving response quality
    \item \textbf{Adaptive granularity:} The topology automatically adjusts to changing demand patterns without manual intervention
\end{itemize}

This semantic organization complements our reputation and consensus mechanisms by ensuring that the most relevant nodes participate in each decision, thereby improving both efficiency and quality of the swarm's collective intelligence.

\section{Enhanced Pairwise Ranking Consensus Mechanism}

Our consensus mechanism represents a fundamental innovation in distributed decision-making, moving beyond simple majority voting to leverage the power of pairwise comparisons. This section details the theoretical foundations, algorithmic implementation, and practical optimizations that enable our system to achieve superior accuracy while maintaining computational efficiency.

\subsection{Theoretical Foundations}

\subsubsection{Cognitive Basis of Pairwise Comparison}

The superiority of pairwise comparison over absolute rating stems from fundamental principles in cognitive psychology and decision theory. Thurstone's Law of Comparative Judgment, formulated in 1927, establishes that humans make more consistent and accurate relative judgments than absolute ones \citep{thurstone1927law}. This principle extends to artificial intelligence systems, where the complexity of assigning absolute quality scores is reduced to the simpler task of determining relative preference.

When evaluating two responses side-by-side, both human and AI evaluators can directly compare specific attributes: factual accuracy, logical coherence, completeness, and stylistic quality. This direct comparison eliminates the need for maintaining consistent internal scales across different evaluation sessions. Furthermore, pairwise comparison naturally handles the context-dependency of quality—a response that would score poorly in absolute terms might be the better option when compared to an even worse alternative.

Recent neuroscience research provides additional support, showing that the brain uses relative coding schemes for value representation, with neurons encoding differences rather than absolute magnitudes \citep{louie2011neural}. This biological evidence suggests that pairwise comparison aligns with fundamental information processing mechanisms, potentially explaining its effectiveness across both natural and artificial intelligence systems.

\subsubsection{Bradley-Terry Model}

The Bradley-Terry model provides an elegant mathematical framework for converting pairwise comparisons into global rankings. For items $i$ and $j$ with latent quality scores $\pi_i$ and $\pi_j$, the probability of preferring $i$ over $j$ follows:

\begin{equation}
P(i \succ j) = \frac{\pi_i}{\pi_i + \pi_j} = \frac{1}{1 + e^{-(\log \pi_i - \log \pi_j)}}
\end{equation}

This logistic formulation has several desirable properties. First, it naturally ensures probabilities remain in $[0,1]$. Second, the model exhibits transitivity in expectation—if $i$ is preferred to $j$ and $j$ to $k$, then $i$ is likely preferred to $k$, though not deterministically. Third, the log-odds form $\log[P(i \succ j)/P(j \succ i)] = \log \pi_i - \log \pi_j$ reveals that differences in log-quality scores directly translate to log-odds ratios, providing interpretable scaling.

The maximum likelihood estimation for Bradley-Terry parameters given observed comparisons has well-established statistical properties. Under mild regularity conditions, the MLE is consistent, asymptotically normal, and efficient. As an extension of the classical theory, Sun et al. demonstrate that Bradley–Terry models with deep network parameterizations can learn complex quality functions \citep{sun2024rethinking}.

\subsubsection{Information-Theoretic Optimality}

From an information-theoretic perspective, pairwise comparisons provide near-optimal information extraction about relative quality. Each comparison can be viewed as a single bit of information (though probabilistic rather than deterministic). The question becomes: how many bits are needed to reliably rank $n$ items?

Classical results from sorting theory establish that $\Omega(n \log n)$ comparisons are necessary and sufficient for deterministic sorting. In a probabilistic setting with noisy comparisons, the Bradley-Terry model achieves a similar efficiency. With $O(n \log n)$ random pairwise comparisons, the model can recover the true ranking with high probability, assuming sufficient separation between quality scores. Our choice of up to $3N$ comparisons per node, yielding $O(N^2)$ total comparisons for $N$ nodes, provides redundancy that enables robust ranking even with Byzantine nodes and noisy evaluations.

Although not a part of a tech report, our team utilizes a more efficient pair sampling approach, yielding \textit{linear} complexity on average.

\subsection{Distributed Comparison Generation}

The process of generating and distributing pairwise comparisons across the node swarm requires careful design to ensure coverage, randomness, and resistance to manipulation.

\subsubsection{Cryptographically Secure Randomization}

Each node generates its assigned comparisons using a cryptographically secure pseudo-random number generator (CSPRNG) seeded with blockchain state. Specifically, we use the hash of the most recent block combined with the node's identifier:

\begin{equation}
\text{seed}_i = \text{SHA256}(\text{state\_hash} \| \text{node\_id}_i)
\end{equation}

This approach ensures that comparison assignments are unpredictable before state finalization, preventing nodes from strategically positioning their responses, yet deterministic afterward, enabling verification that nodes completed assigned comparisons. The use of blockchain-based randomness also provides consensus on random values without requiring additional communication rounds.

\subsubsection{Balanced Comparison Distribution}

The random assignment process must ensure adequate coverage while avoiding excessive redundancy. We employ a modified coupon collector approach where each node independently samples comparison pairs, with the total number chosen to ensure high probability of complete coverage.

For $N$ responses and $M$ nodes each generating $3N$ comparisons, the probability that any specific pair $(i,j)$ is never compared is:

\begin{equation}
P(\text{pair missed}) = \left(1 - \frac{1}{\binom{N}{2}}\right)^{3NM} \approx e^{-\frac{6M}{N-1}}
\end{equation}

With typical values of $M \approx N$, this probability becomes negligibly small, ensuring comprehensive evaluation coverage.

\subsubsection{Exclusion of Self-Comparisons}

Nodes cannot rank their own responses to prevent self-promotion bias. This exclusion is enforced cryptographically—each response is signed by its generating node, and the ranking engine verifies that no self-comparisons are included in submitted rankings. Attempts to include self-comparisons result in immediate rejection and reputation penalties.

This exclusion creates an interesting game-theoretic dynamic: nodes have incentives to generate high-quality responses (to win consensus), but cannot directly influence their own ranking. This separation of generation and self-evaluation encourages honest participation while maintaining competitive incentives for quality.

\subsection{Multi-Token Reasoning Chains}

The requirement for explicit reasoning chains represents a crucial innovation that distinguishes our approach from single-scalar reward models. Each pairwise comparison must be accompanied by a structured reasoning chain that addresses specific quality dimensions. The example template enforces systematic evaluation and smallest amount of tokens produced.

\begingroup
\small
\begin{verbatim}
RANKING_TEMPLATE = """
    Select the best one of two following solutions. WRITE up to 3 the most important
    and UNIQUE mistakes/contradictions for a particular response (Like: Solution N
    lacks ...) with EXTREMELY concise and exact notes, THEN end output with best
    solution overall index (1 or 2) on the new line (Only number, nothing else).
    If there are no obvious winner solution (or they are the same) respond with
    'Uncertain?'. Make sure that solution is finished and does not abruptly cut off.
    Carefully check solution's consistency with the requirements and comments
    with the problem definition.
    
    ######Problem######: {R}
    ######1######. {A}
    ######2######. {B}
    ######Decision######:
"""
\end{verbatim}
\endgroup

This structured approach ensures a comprehensive evaluation rather than quick judgments based on surface characteristics. The requirement to explicitly identify strengths and weaknesses in each response forces a deeper analysis and reduces cognitive biases. In practical conditions, it is beneficial to use different prompts with different focuses to steer the ranking into a wider decision space.

\subsubsection{Cognitive Load and Quality Trade-offs}

Requiring detailed reasoning increases computational cost, as generating 50-100 tokens of explanation requires significantly more processing than producing a single classification token. However, our experiments demonstrate that this investment yields substantial returns in ranking accuracy.

The cognitive load of producing explanations engages what psychologists term "System 2" thinking: deliberate, analytical processing that reduces errors compared to intuitive "System 1" judgments \citep{kahneman2011thinking}. In the context of language models, this translates to more careful attention to details and systematic consideration of multiple factors rather than pattern-matching based on superficial features.

\subsubsection{Reasoning Chain Analysis}

The generated reasoning chains provide valuable metadata beyond their primary purpose of improving ranking accuracy. Statistical analysis of reasoning patterns reveals:

\begin{itemize}
    \item \textbf{Systematic biases:} Nodes that consistently favor certain types of responses (e.g., longer answers, technical jargon) can be identified and calibrated
    \item \textbf{Expertise indicators:} Nodes demonstrating sophisticated reasoning in specific domains can be preferentially assigned related queries
    \item \textbf{Quality predictors:} Characteristics of high-quality reasoning (specificity, evidence citation, logical consistency) correlate with ranking accuracy
    \item \textbf{Manipulation detection:} Templated or repetitive reasoning suggests lazy evaluation or attempted gaming
\end{itemize}

This analysis enables continuous system improvement through identification of effective evaluation strategies and detection of problematic patterns.

\subsection{Bradley-Terry Aggregation Algorithm}

The aggregation of pairwise comparisons into global rankings requires additional optimization techniques to handle the scale and noise inherent in distributed evaluation.

Given comparison outcomes where response $i$ was preferred to response $j$ exactly $w_{ij}$ times out of $n_{ij}$ comparisons, the log-likelihood function is:

\begin{equation}
\ell(\boldsymbol{\pi}) = \sum_{i<j} \left[ w_{ij} \log \frac{\pi_i}{\pi_i + \pi_j} + (n_{ij} - w_{ij}) \log \frac{\pi_j}{\pi_i + \pi_j} \right]
\end{equation}

Maximizing this likelihood yields quality scores that best explain the observed comparisons. The optimization problem is convex in the log-parameters $\boldsymbol{\theta} = \log \boldsymbol{\pi}$, ensuring convergence to a global optimum.

\subsection{Reputation-Weighted Consensus}

The integration of reputation weights into the consensus mechanism creates a meritocratic system where influence is earned through consistent quality contributions.

\begin{figure}[h]
\centering
\begin{tikzpicture}
    \begin{axis}[
        width=10cm,
        height=6cm,
        xlabel={Consensus Rounds},
        ylabel={Reputation Score},
        legend pos=north west,
        grid=major,
        ymin=0, ymax=1,
        xmin=0, xmax=100
    ]
    
    \addplot[color=blue, thick] coordinates {
        (0,0.5) (10,0.55) (20,0.62) (30,0.70) (40,0.78) 
        (50,0.84) (60,0.88) (70,0.91) (80,0.93) (90,0.94) (100,0.95)
    };
    \addlegendentry{High Performer}
    
    \addplot[color=green, thick] coordinates {
        (0,0.5) (10,0.52) (20,0.54) (30,0.56) (40,0.57) 
        (50,0.58) (60,0.59) (70,0.595) (80,0.60) (90,0.60) (100,0.60)
    };
    \addlegendentry{Average Performer}
    
    \addplot[color=red, thick] coordinates {
        (0,0.5) (10,0.48) (20,0.44) (30,0.38) (40,0.31) 
        (50,0.24) (60,0.18) (70,0.14) (80,0.11) (90,0.09) (100,0.08)
    };
    \addlegendentry{Poor Performer}
    
    \addplot[color=orange, thick, dashed] coordinates {
        (0,0.5) (10,0.45) (20,0.38) (30,0.30) (40,0.28) 
        (50,0.32) (60,0.40) (70,0.48) (80,0.54) (90,0.58) (100,0.61)
    };
    \addlegendentry{Recovery Pattern}
    
    \end{axis}
\end{tikzpicture}
\caption{Evolution of node reputation scores over time showing different performance patterns}
\label{fig:reputation_evolution}
\end{figure}
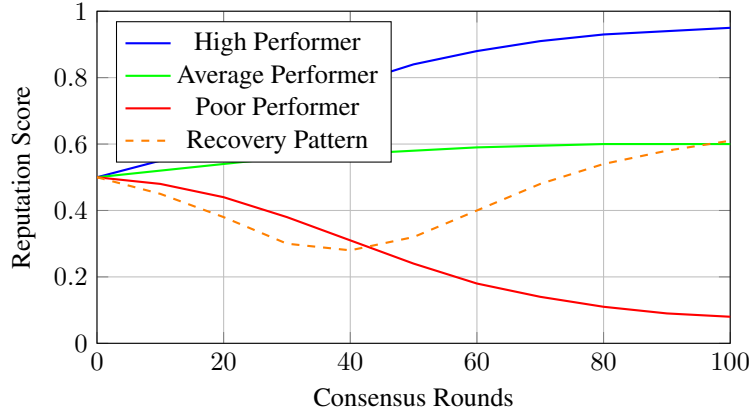

\subsubsection{Reputation Score Computation}

Node reputation evolves based on two primary metrics: ranking accuracy (how well their comparisons align with final consensus) and response quality (how often their generated responses win consensus). The combined reputation score is:

\begin{equation}
R_i = \alpha \cdot R_{\text{ranking},i} + (1-\alpha) \cdot R_{\text{generation},i}
\end{equation}

where $\alpha$ balances the importance of evaluation versus generation capability.

Ranking accuracy is measured using Kendall's tau correlation between an node's comparisons and the final consensus ranking with exponential moving average applied.

\begin{equation}
R_{\text{ranking},i}^{(t+1)} 
= \beta R_{\text{ranking},i}^{(t)} 
+ (1-\beta) \cdot 
\frac{1}{|\mathcal{C}_i|} 
\sum_{(j,k)\in\mathcal{C}_i} 
\mathds{1}\!\bigl[\text{sign}(c_{i,jk}) = \text{sign}(\pi_j - \pi_k)\bigr]
\end{equation}

Generation quality uses an exponential moving average of consensus wins:

\begin{equation}
R_{\text{generation},i}^{(t+1)} = \beta R_{\text{generation},i}^{(t)} + (1-\beta) \cdot \mathds{1}[i = \text{winner}_t]
\end{equation}

\subsubsection{Weighted Bradley-Terry Model}

Reputation weights are incorporated directly into the Bradley-Terry optimization by scaling each comparison's contribution:

\begin{equation}
\ell_{\text{weighted}}(\boldsymbol{\pi}) = \sum_{a \in \text{Nodes}} R_a \sum_{(i,j) \in \mathcal{C}_a} \left[ y_{ij}^{(a)} \log \frac{\pi_i}{\pi_i + \pi_j} + (1-y_{ij}^{(a)}) \log \frac{\pi_j}{\pi_i + \pi_j} \right]
\end{equation}

This weighting ensures that comparisons from high-reputation nodes have greater influence on final rankings while still incorporating information from all participants.

\subsubsection{Dynamic Reputation Updates}

Reputation updates occur after each consensus round, creating a dynamic system that responds to performance changes:

\begin{equation}
R_i^{(t+1)} = \begin{cases}
\min(R_{\max}, R_i^{(t)} + \Delta_{\text{up}}) & \text{if performance exceeds threshold} \\
\max(R_{\min}, R_i^{(t)} - \Delta_{\text{down}}) & \text{if performance below threshold} \\
R_i^{(t)} \cdot (1 - \delta) & \text{if inactive}
\end{cases}
\end{equation}

The system includes caps on maximum reputation to prevent excessive concentration of power, and decay for inactive nodes to ensure continued participation.

\section{Compute Stake Mechanism and Sybil Defense}

The prevention of Sybil attacks—where adversaries create multiple fake identities to gain disproportionate influence—represents a fundamental challenge in decentralized systems. Our compute stake mechanism introduces a novel approach that requires demonstration of computational capability and quality rather than economic investment, creating natural barriers against identity multiplication while maintaining openness to legitimate participants.

\subsection{Theoretical Framework}

\subsubsection{Limitations of Traditional Approaches}

Classical Sybil defense mechanisms rely on making identity creation expensive through resource expenditure. Proof-of-Work requires solving computational puzzles, consuming energy proportional to influence gained \citep{nakamoto2008bitcoin}. Proof-of-Stake demands locking economic value, with influence proportional to staked assets \citep{kiayias2017ouroboros}. Both approaches suffer from fundamental limitations:

\begin{itemize}
    \item \textbf{Wealth concentration:} Economic barriers favor participants with greater capital resources, contradicting decentralization goals
    \item \textbf{Resource waste:} PoW consumes enormous energy without producing useful work
    \item \textbf{Capital inefficiency:} PoS locks liquidity that could be productively deployed elsewhere
    \item \textbf{Plutocracy risks:} Both systems tend toward centralization as resources concentrate
\end{itemize}

Our compute stake mechanism addresses these limitations by requiring proof of capability, demonstrated competence in performing useful work for the swarm, rather than proof of resource expenditure.

\subsubsection{Computational Proof-of-Capability}

The key insight is that creating a competent AI node requires both computational investment in either training or inference infrastructure as well as intelligent selection and calibration to provide useful capability the swarm. That greatly limits scaling, compared to resource proofs that can be easily outsourced or pooled.

Formally, let $C(q)$ be the computational cost of correctly answering query $q$, and $Q_s = \{q_1, ..., q_n\}$ be a test requests that are part of the selected node's specialization $S$ (by assigning semantic embeddings). An node must demonstrate:

\begin{equation}
\text{Capability} = \sum_{i=1}^{n} \mathds{1}[\text{correct}(q_i)] \cdot C(q_i) \geq \tau
\end{equation}

where $\tau$ is the minimum capability threshold. This differs from PoW where computation produces no useful output; here, the computation directly demonstrates the capability needed for system participation and used later in datasets or as a free inference if it surpasses specified quality level. Moreover, after achieving the threshold $\tau$ nodes maintain their reputation in rewarded requests, thus doing only useful work for the swarm. 

\subsection{Test Requests Design and Implementation}

\subsubsection{Comprehensive Domain Coverage}

The test suite spans multiple semantic domains to ensure a well-rounded capability assessment. Node can pick any semantic domains that its expert in. Each domain tests different aspects of intelligence that are valuable for consensus participation.

\paragraph{Mathematical Reasoning:} Problems require multi-step logical inference, symbolic manipulation, and proof construction. Examples include solving systems of equations, proving geometric theorems, and optimizing complex functions. These tests verify logical consistency and systematic reasoning capabilities essential for quality evaluation.

\paragraph{Scientific Analysis:} Questions span physics, chemistry, biology, and interdisciplinary topics requiring domain knowledge and scientific reasoning. Tasks include interpreting experimental data, formulating hypotheses, and identifying flaws in scientific arguments. This ensures nodes can evaluate technical content accurately.

\paragraph{Code Generation and Analysis:} Programming challenges across multiple languages test both synthesis and analysis capabilities. Tasks range from implementing algorithms to identifying security vulnerabilities and optimizing performance. These skills are crucial for evaluating code-based responses common in many queries.

\paragraph{Natural Language Understanding:} Complex comprehension tasks, ambiguity resolution, and logical reasoning over text evaluate language capabilities. This includes understanding implicit information, recognizing argumentative fallacies, and synthesizing information from multiple sources.

\paragraph{Creative Problem-Solving:} Open-ended challenges requiring innovative approaches test flexibility and creativity. These problems have multiple valid solutions, evaluating an node's ability to generate novel approaches rather than retrieve memorized answers.

\subsubsection{Dynamic Test Generation}

Static test suites are vulnerable to gaming through memorization or specialized preparation. Our system employs dynamic test generation to maintain security:

\begin{equation}
q_{\text{test}} = T(q_{\text{template}}, \text{seed}_{\text{random}})
\end{equation}

where $T$ is a transformation function that modifies template problems using random seeds. Transformations include:

\begin{itemize}
    \item Numerical parameter variation within reasonable ranges
    \item Semantic paraphrasing while preserving problem structure
    \item Combining multiple simpler problems into compound challenges
    \item Contextual embedding in different domains or scenarios
\end{itemize}

The test generator maintains difficulty calibration through statistical analysis of pass rates, ensuring consistent challenge levels despite randomization. 

Another source of tests are free requests. Free requests sponsored by the network are then used as a test case if real node struggle on average with it's completion. In such way after initial seeding tests with test suites, the network growth the test corpus organically.  

\subsection{Sybil Attack Models and Defense}

\subsubsection{Attack Objectives}

A Sybil attack in our consensus system has two primary objectives:
\begin{enumerate}
\item \textbf{Consensus Manipulation}: Force the system to select attacker-controlled responses as winners regardless of actual quality
\item \textbf{Reputation Hijacking}: Build artificial reputation for Sybil nodes while suppressing legitimate participants
\end{enumerate}

The attacker creates $k$ colluding identities $S = \{s_1, s_2, \ldots, s_k\}$ that coordinate to:
\begin{itemize}
\item Generate low-quality responses using cheap models or random noise
\item Rank each other's responses highly (collusion voting)
\item Rank legitimate nodes' responses poorly
\item Maximize collective reward extraction
\end{itemize}

\subsubsection{Defensive Mechanisms}

\paragraph{Collusion Detection via Voting Pattern Analysis}

We detect collusion by analyzing mutual support patterns. For each pair of nodes $(i,j)$, we track:
\begin{equation}
c_{ij}(t) = \frac{\text{\# of times } i \text{ ranked } j \text{ in top half}}{\text{\# of rounds both participated}}
\end{equation}

Under random assignment, $\mathbb{E}[c_{ij}] \approx \frac{N}{2 \cdot (N-1)}$ for $N$ participants. Nodes with $c_{ij} \gg \mathbb{E}[c_{ij}]$ indicate potential collusion.

The reputation weight is exponentially diminished for suspected colluders:
\begin{equation}
w_{ij}^{\text{adjusted}} = w_j \cdot \exp\left(-\lambda \cdot \max(0, c_{ij} - \tau_{\text{collusion}})\right)
\end{equation}

where $\lambda$ is the penalty coefficient (typically 10-20) and $\tau_{\text{collusion}}$ is the threshold (typically $1.2\mathbb{E}[c_{ij}]$).

\paragraph{Performance-Based Reputation Slashing}

Nodes with consistently poor ranking quality face reputation penalties. When reputation drops below threshold level $Q$ its automatically slashed to $0$ requiring the node to redo full test recalibration against the ground truth verifier. 

\paragraph{Abnormal Participation Filtering}

Rounds with unusual participation patterns receive reduced weight in reputation calculations:
\begin{equation}
w_{\text{round}} = \exp\left(-\gamma \cdot \left|\log\frac{n_{\text{actual}}}{\bar{n}}\right|\right)
\end{equation}

where $n_{\text{actual}}$ is the round's participant count, $\bar{n}$ is the historical average, and $\gamma$ controls sensitivity (typically 1-2).

\subsubsection{Economic Analysis with Defenses}

The expected reward for a Sybil attacker with $k$ identities under our defensive mechanisms:
\begin{equation}
\text{Revenue}_{\text{Sybil}}(k) = \sum_{t=1}^{T} w_{\text{round}}^{(t)} \cdot \sum_{i=1}^{k} \frac{R_i^{(t)} \cdot \prod_{j \neq i} \exp(-\lambda c_{ij})}{\sum_{m \in \text{All}} R_m^{(t)}} \cdot \text{Reward}^{(t)}
\end{equation}

The collusion penalty term $\prod_{j \neq i} \exp(-\lambda c_{ij})$ rapidly diminishes rewards as mutual support becomes apparent.

\paragraph{Cost Structure}

Creating and maintaining $k$ Sybil identities incurs:
\begin{align}
\text{Cost}_{\text{entry}}(k) &= k \cdot C_{\text{test}} \cdot n_{\text{tests}} \\
\text{Cost}_{\text{operation}}(k,t) &= k \cdot C_{\text{inference}} \cdot n_{\text{rounds}}(t) \\
\text{Cost}_{\text{slashing}}(k,t) &= k \cdot R_{\text{initial}} \cdot \alpha_{\text{slash}} \cdot f_{\text{detected}}(t)
\end{align}

where $f_{\text{detected}}(t)$ is the fraction of rounds where collusion is detected and penalized.

\paragraph{Break-Even Analysis}

For profitability:
\begin{equation}
\sum_{t=1}^{T} \text{Revenue}_{\text{Sybil}}(k) > \text{Cost}_{\text{entry}}(k) + \sum_{t=1}^{T} \text{Cost}_{\text{operation}}(k,t) + \sum_{t=1}^{T} \text{Cost}_{\text{slashing}}(k,t)
\end{equation}

Our simulations show that this inequality cannot be satisfied for $\lambda > 10$ and $k > 1$ even with perfect collusion coordination if $\text{Cost}_{\text{entry}} \gg \text{Cost}_{\text{operation}}$ is achieved with a test calibration period long enough.

\subsection{Reputation Dynamics and Long-term Incentives}

\subsubsection{Reputation as Non-Transferable Asset}

Unlike economic stake that can be sold or transferred, reputation in our system is intrinsically tied to node identity through cryptographic commitments:

\begin{equation}
R_i = H(\text{history}_i, \text{pubkey}_i)
\end{equation}

This non-transferability creates several desirable properties:

\begin{itemize}
    \item \textbf{No reputation markets:} Reputation cannot be bought, only earned through performance
    \item \textbf{Skin in the game:} Nodes cannot exit without losing accumulated reputation value
    \item \textbf{Long-term alignment:} Building reputation requires sustained quality contributions
\end{itemize}

\subsubsection{Evolutionary Dynamics}

The reputation system creates evolutionary pressure toward quality:

\begin{itemize}
    \item High-quality nodes earn more and gain influence
    \item Poor performers are marginalized and eventually excluded
    \item The population evolves toward higher average capability
    \item System performance improves over time without explicit optimization
\end{itemize}

This natural selection process makes the system antifragile—attacks and failures strengthen rather than weaken the system by accelerating the removal of weak nodes.

\section{Evaluation}

Our comprehensive experimental evaluation demonstrates the effectiveness of the Fortytwo Protocol across multiple dimensions: accuracy on diverse benchmarks, robustness to adversarial conditions, scalability with swarm size, and economic viability. We present detailed results with statistical analysis, ablation studies isolating component contributions, and comparison with state-of-the-art alternatives.

\subsection{Experimental Methodology}

\subsubsection{Benchmark Selection}

We evaluate our system on six challenging benchmarks selected to test different aspects of AI capability:

\begin{itemize}
    
    \item \textbf{GPQA Diamond:} 198 graduate-level science questions where human experts achieve only 65\% accuracy, testing cutting-edge knowledge in physics, chemistry, and biology \citep{rein2023gpqa}
    
    \item \textbf{LiveCodeBench:} v5 subset of 315 contamination-free coding problems updated monthly to prevent overfitting, covering algorithms, data structures, and real-world programming challenges \citep{jain2024livecodebench}
    
    \item \textbf{HLE (Humanity Last Exam):} 2,500 problems at the frontier of human knowledge, designed to be the final closed-ended academic benchmark of its kind with broad subject coverage. HLE consists of dozens of subjects, including mathematics, humanities, and the natural sciences.\citep{hle2025}
    
    \item \textbf{MATH-500:} 500 competition mathematics problems sampled from MATH dataset, spanning topics like probability, algebra, trigonometry, and geometry. The questions are designed to test a model’s ability to apply mathematical principles, execute complex calculations, and communicate solutions clearly. \citep{hendrycks2021measuring}
    
    \item \textbf{AIME 2024-2025:} 60 problems from recent American Invitational Mathematics Examinations, representing elite competition-level mathematical reasoning
\end{itemize}

\subsubsection{Swarm Configuration}

We test the most powerful configuration while balancing it with practical deployment constraints:

\paragraph{Test Configuration (35 nodes):}
\begin{itemize}
    \item 12× GLM 4.5 (temperatures: 0.0, 0.7)
    \item 17× GPT-OSS 120B (temperatures: 0.0, 0.7)
    \item 7× QWEN3 235B (Reasoning) (temperatures: 0.0, 0.7)
    \item Target: High-end hardware (40GB+ VRAM)
\end{itemize}

Temperature variation ensures response diversity while maintaining quality. Zero-temperature instances provide deterministic baselines, whereas higher temperatures encourage creative solutions.

All experiments (except MMLU Pro and HLE benchmarked with a single run) use 5-fold cross-validation with different random seeds.

\subsection{Main Results}
\subsubsection{Benchmark Performance}

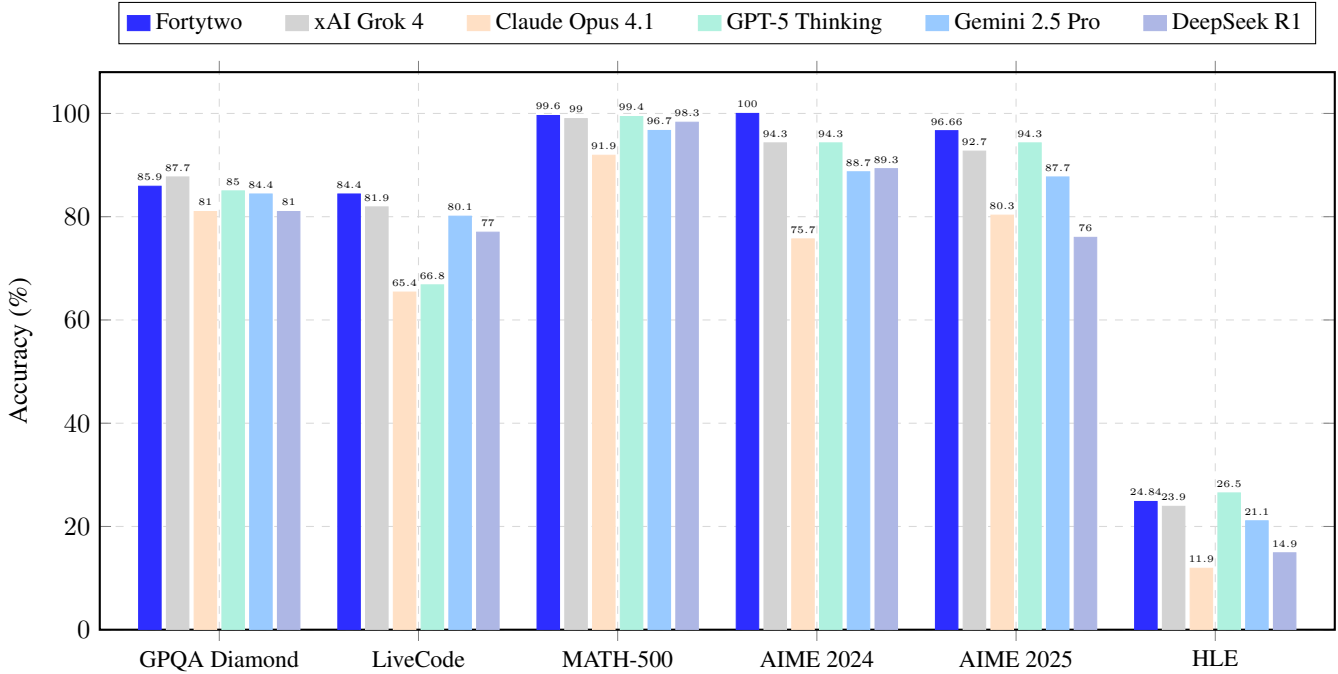
\begin{figure}[htbp]
\centering

\begin{tikzpicture}
\definecolor{fortytwo}{RGB}{45, 45, 255}
\definecolor{grok}{RGB}{211, 211, 211}
\definecolor{opus}{RGB}{255, 224, 197}
\definecolor{gpt5}{RGB}{175, 241, 223}
\definecolor{gemini}{RGB}{153, 202, 255}
\definecolor{deepseek}{RGB}{174, 183, 229}

\begin{axis}[
    ybar,
    bar width=8.5pt,
    width=18cm,
    height=9cm,
    ylabel={Accuracy (\%)},
    ylabel style={font=\normalsize},
    symbolic x coords={GPQA Diamond, LiveCode, MATH-500, AIME 2024, AIME 2025, HLE}, 
    xtick=data,
    xticklabel style={font=\small},
    ymin=0, ymax=108,
    ytick={0,20,40,60,80,100},
    legend style={
        at={(0.5,1.05)},
        anchor=south,
        legend columns=7,
        font=\small,
        /tikz/every even column/.append style={column sep=0.5cm},
    },
    legend image code/.code={
            \fill[#1] (0cm,-0.1cm) rectangle (0.3cm,0.1cm);
    },
    nodes near coords,
    nodes near coords style={font=\fontsize{3.8}{3.8}\selectfont, anchor=south, yshift=-1.2pt},
    enlarge x limits=0.12,
    grid=major,
    grid style={dashed,gray!30},
    ymajorgrids=true,
    axis line style={thick}
]

\addplot[fill=fortytwo, draw=fortytwo] coordinates {
    (GPQA Diamond, 85.90)
    (LiveCode, 84.40)
    (MATH-500, 99.60)
    (HLE, 24.84)
    (AIME 2024, 100.0)
    (AIME 2025, 96.66)
};
\addlegendentry{Fortytwo}

\addplot[fill=grok, draw=grok] coordinates {
    (GPQA Diamond, 87.70)
    (LiveCode, 81.90)
    (MATH-500, 99.00)
    (HLE, 23.90)
    (AIME 2024, 94.30)
    (AIME 2025, 92.70)
};
\addlegendentry{xAI Grok 4}

\addplot[fill=opus, draw=opus] coordinates {
    (GPQA Diamond, 81.00)
    (LiveCode, 65.40)
    (MATH-500, 91.90)
    (HLE, 11.90)
    (AIME 2024, 75.70)
    (AIME 2025, 80.30)
};
\addlegendentry{Claude Opus 4.1}

\addplot[fill=gpt5, draw=gpt5] coordinates {
    (GPQA Diamond, 85.00)
    (LiveCode, 66.80)
    (MATH-500, 99.40)
    (HLE, 26.50)
    (AIME 2024, 94.30)
    (AIME 2025, 94.30)
};
\addlegendentry{GPT-5 Thinking}

\addplot[fill=gemini, draw=gemini] coordinates {
    (GPQA Diamond, 84.40)
    (LiveCode, 80.10)
    (MATH-500, 96.70)
    (HLE, 21.10)
    (AIME 2024, 88.70)
    (AIME 2025, 87.70)
};
\addlegendentry{Gemini 2.5 Pro}

\addplot[fill=deepseek, draw=deepseek] coordinates {
    (GPQA Diamond, 81.00)
    (LiveCode, 77.00)
    (MATH-500, 98.30)
    (HLE, 14.90)
    (AIME 2024, 89.30)
    (AIME 2025, 76.0)
};
\addlegendentry{DeepSeek R1}

\end{axis}
\end{tikzpicture}

\caption{Benchmark performance comparison across models. Fortytwo achieves state-of-the-art results on LiveCode (84.4\%), MATH-500 (99.6\%), AIME 2024 (100\%), and AIME 2025 (96.66\%), demonstrating superior performance on challenging reasoning and coding benchmarks.}
\label{fig:benchmark_performance}
\end{figure}

\begin{itemize}
        
    \item \textbf{Competitive with SOTA:} While individual models like xAI Grok 4 achieve higher scores on some benchmarks, our system provides more consistent performance across diverse domains    
\end{itemize}

\subsection{Ablation Studies}

\subsubsection{Component Contribution Analysis}

We systematically remove or modify system components to assess their individual contributions:

\begin{table}[h]
\centering
\caption{Ablation study results on GPQA Diamond (\% accuracy)}
\label{tab:ablation}
\begin{tabular}{lcc}
\toprule
\textbf{Configuration} & \textbf{Accuracy} & \textbf{$\Delta$ from Full} \\
\midrule
Full System & 84.9 & -- \\
\midrule
Without ranking weighting & 83.2 & -1.7 \\
With single model ranking & 82.4 & -2.5 \\
Without reasoning chains & 79.6 & -5.3 \\
Single temperature (no diversity) & 75.8 & -10.1 \\
\bottomrule
\end{tabular}
\end{table}

Critical insights from ablation analysis:

\begin{itemize}
    \item \textbf{Reasoning chains are crucial:} Removing multi-token reasoning requirements causes large performance drop (-5.3\%), confirming our hypothesis about deliberative evaluation

    \item \textbf{Temperature sampling is necessary for exploration:} Removing temperature sampling, caused the largest performance drop, confirming that exploration is the base of the swarm inference approach
    
    \item \textbf{Ranking with different models improves quality:} Using single model for the ranking, as well as adjusting their ranking weight based on performance significantly affects the quality, thus showing the importance of unbiased judging of the answers
    
    \item \textbf{All components contribute:} Every system component provides measurable value, validating our holistic design approach
\end{itemize}

\subsubsection{Swarm Size Scaling}

We examine how performance scales with the number of nodes in GPQA Diamond:

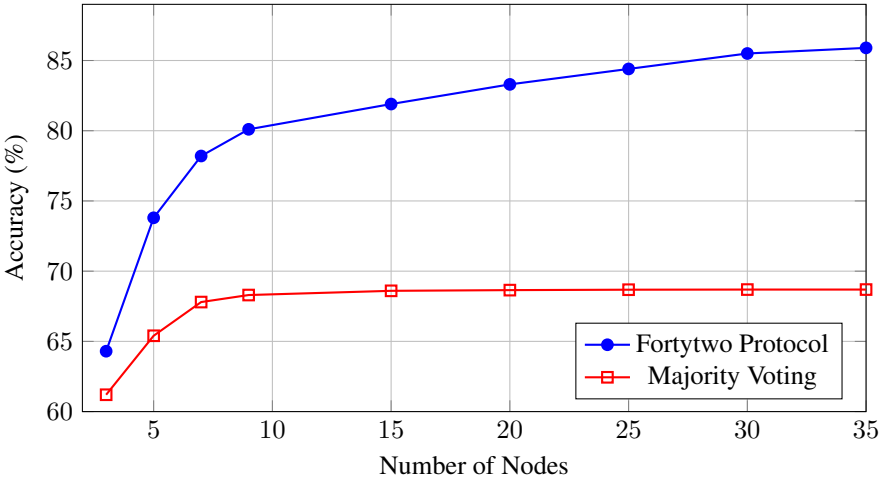
\begin{figure}[h]
\centering
\begin{tikzpicture}
    \begin{axis}[
        width=12cm,
        height=7cm,
        xlabel={Number of Nodes},
        ylabel={Accuracy (\%)},
        legend pos=south east,
        grid=major,
        xmin=2, xmax=35,
        ymin=60, ymax=89
    ]
    
    \addplot[color=blue, thick, mark=*] coordinates {
        (3,64.3) (5,73.8) (7,78.2) (9,80.1) (15,81.9) 
        (20,83.3) (25,84.4) (30,85.5) (35,85.9)
    };
    \addlegendentry{Fortytwo Protocol}
    
    \addplot[color=red, thick, mark=square] coordinates {
        (3,61.2) (5,65.4) (7,67.8) (9,68.3) (15,68.6) 
        (20,68.65) (25,68.68) (30,68.69) (35,68.69)
    };
    \addlegendentry{Majority Voting}
    
    \end{axis}
\end{tikzpicture}
\caption{Performance scaling with swarm size showing rapid improvement and convergence}
\label{fig:swarm_scaling}
\end{figure}

Key observations from scaling analysis:

\begin{itemize}
    \item \textbf{Rapid initial improvement:} Performance increases sharply from 3 to 7 nodes, demonstrating the value of diverse perspectives
    
    \item \textbf{Efficient convergence:} Performance plateaus around 30 nodes, suggesting that moderate-sized swarms capture most benefits, although open question remains how this scaling would behave with more diverse mixture of models
    
    \item \textbf{Consistent advantage:} Fortytwo maintains superiority over majority voting at all swarm sizes
    
    \item \textbf{Theoretical alignment:} Empirical results closely match theoretical predictions based on information aggregation models
\end{itemize}

\subsection{Robustness Analysis}
\subsubsection{Reasoning Resilience: Robustness to Extraneous Information}

A critical challenge in real-world AI deployment is maintaining performance when faced with noisy, verbose, or deliberately misleading inputs. To evaluate this capability, we analyze performance on GPQA Diamond with and without extraneous information—irrelevant details deliberately inserted to test whether models truly understand the problem or merely pattern-match on superficial features.

\paragraph{GPQA Diamond with Extraneous Information.} This benchmark variant augments standard GPQA Diamond questions with irrelevant contextual information designed to test reasoning robustness. For example, a chemistry question might include unrelated personal anecdotes or tangential facts that a model must filter out to identify the core problem.

\textbf{Example prompt comparison:}

\begin{tcolorbox}[colback=gray!10, colframe=gray!50, title=Standard GPQA Diamond Question]
\small
\texttt{Answer the following question. The last line of your response should be in the following format: 'Answer: A/B/C/D'.}

\vspace{0.2cm}

\texttt{Acetic acid is treated with <...> how many distinct hydrogen signals will be observable in the 1H NMR spectrum of 4? <...>}

\vspace{0.1cm}

\texttt{A) 5\\
B) 10\\
C) 12\\
D) 8}
\end{tcolorbox}

\begin{tcolorbox}[colback=blue!5, colframe=blue!50, title=GPQA Diamond with Extraneous Information]
\small
\texttt{Answer the following question. The last line of your response should be in the following format: 'Answer: \textcolor{blue}{<LETTER>}'.}

\vspace{0.2cm}

\textcolor{blue}{\texttt{Also, some nonrelevant message: There is a cat on the roof. Maybe it is hungry?!}}

\vspace{0.2cm}

\texttt{Acetic acid is treated with <...> how many distinct hydrogen signals will be observable in the 1H NMR spectrum of 4? <...>}

\vspace{0.1cm}

\texttt{\textcolor{blue}{Z)} 5\\
\textcolor{blue}{X)} 10\\
\textcolor{blue}{C)} 12\\
\textcolor{blue}{V)} 8}
\end{tcolorbox}

Extraneous information (highlighted in blue) is deliberately designed to be unrelated or confusing to the model, testing whether models can maintain focus on the actual problem or become distracted by irrelevant context.
\begin{figure}[h]
\centering
\begin{tikzpicture}
\definecolor{darkbar}{RGB}{80, 80, 80}
\definecolor{lightbar}{RGB}{200, 200, 200}

\begin{axis}[
    xbar,
    width=14cm,
    height=12cm,
    xlabel={Accuracy (\%)},
    xlabel style={font=\small},
    ytick={1,2,3,4,5,6},
    yticklabels={DeepSeek R1, Claude Opus 4.1, Gemini 2.5 Pro, GPT-5 Thinking, xAI Grok 4, Fortytwo},
    yticklabel style={font=\small, anchor=east},
    xmin=0, xmax=100,
    xtick={65,70,75,80,85,90},
    xticklabel style={font=\small},
    bar width=10pt,
    bar shift=0pt,
    legend style={
        at={(0.5,0.02)},
        anchor=south east,
        font=\small,
    },
    legend image code/.code={
            \fill[#1] (0cm,-0.1cm) rectangle (0.3cm,0.1cm);
    },
    nodes near coords,
    nodes near coords style={font=\footnotesize, anchor=west, xshift=2pt},
    grid=major,
    grid style={dashed,gray!30},
    xmajorgrids=true,
]

\addplot[fill=darkbar, draw=none, bar shift=-5pt, forget plot] coordinates {
    (70.2, 1)
    (74.45, 2)
    (83.2, 3)
    (83.8, 4)
    (79.5, 5)
};

\addplot[fill=darkbar, draw=none, bar shift=-5pt] coordinates {
    (85.76, 6)
};
\addlegendentry{With Extraneous Information}

\addplot[fill=lightbar, draw=none, bar shift=5pt] coordinates {
    (81.0, 1)
    (81.0, 2)
    (84.4, 3)
    (85.0, 4)
    (87.7, 5)
    (85.9, 6)
};
\addlegendentry{Without Extraneous Information}

\end{axis}
\end{tikzpicture}
\caption{Reasoning resilience on GPQA Diamond: Performance comparison with and without extraneous information.}
\label{fig:reasoning_resilience}
\end{figure}
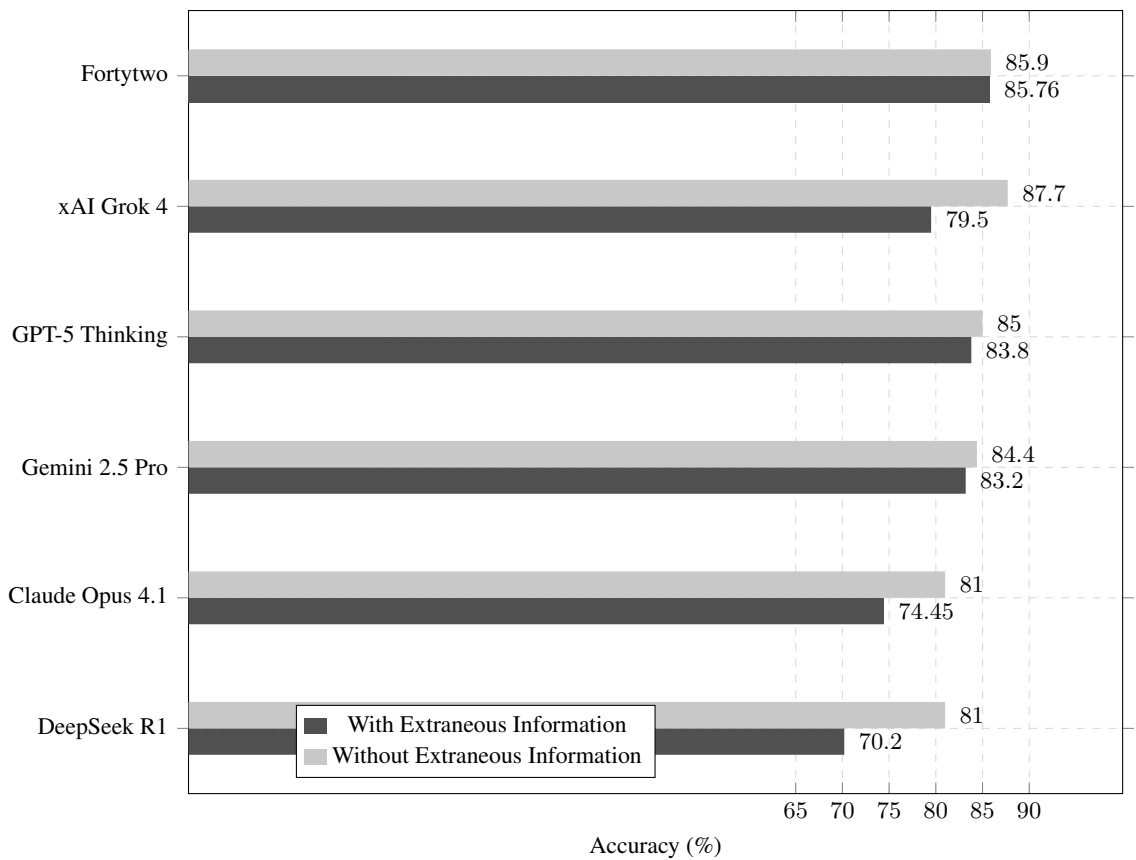

\paragraph{Quantitative Analysis.} Table~\ref{tab:extraneous_degradation} presents the performance degradation when extraneous information is introduced:

\begin{table}[h]
\centering
\caption{Performance degradation with extraneous information (GPQA Diamond)}
\label{tab:extraneous_degradation}
\begin{tabular}{lccc}
\toprule
\textbf{Model} & \textbf{Standard} & \textbf{With Noise} & \textbf{Degradation} \\
\midrule
Fortytwo & 85.90\% & \textbf{85.78}\% & \textbf{-0.12\%} \\
GPT-5 Thinking & 85.00\% & 83.80\% & -1.20\% \\
Gemini 2.5 Pro & 84.40\% & 83.20\% & -1.20\% \\
Claude Opus 4.1 & 81.00\% & 74.45\% & -6.55\% \\
DeepSeek R1 & 81.00\% & 70.20\% & -11.20\% \\
xAI Grok 4 & \textbf{87.70}\% & 79.50\% & -8.20\% \\
\midrule
\textbf{Average (others)} & 83.82\% & 77.62\% & -6.20\% \\
\bottomrule
\end{tabular}
\end{table}

\paragraph{Key Findings.} Our swarm-based approach demonstrates remarkable resilience:

\begin{itemize}
    \item \textbf{Minimal degradation:} Fortytwo experiences only 0.12\% accuracy loss with extraneous information, compared to an average 6.20\% degradation for competing models—representing 52× greater robustness.
    
    \item \textbf{Consensus filtering:} The distributed ranking mechanism effectively filters out noise, as different models are affected differently by extraneous information. When one model is distracted by irrelevant details, others maintain focus on the core problem.
    
    \item \textbf{Reasoning depth:} While single models like Grok 4 can achieve higher baseline accuracy (87.7\%), they suffer severe degradation (-8.20\%) when noise is introduced. Fortytwo's stability indicates robust reasoning that penetrates beyond superficial patterns.
    
    \item \textbf{Real-world applicability:} Free-form user prompts rarely follow clean, structured formats. Our system's resilience to extraneous information translates directly to superior performance in production environments where users provide verbose, context-rich, or poorly formatted queries.

    \item \textbf{Prompt injection resistance:} The peer-ranked consensus inherently mitigates prompt injection attacks, as malicious or misleading instructions must influence a majority of independent nodes simultaneously to affect the final outcome. Disagreement among nodes exposed to injected content leads to low consensus weighting for compromised responses, effectively neutralizing adversarial influence.
\end{itemize}

\paragraph{Mechanistic Explanation.} The swarm's robustness emerges from three key mechanisms:

\begin{enumerate}
    \item \textbf{Diversity of attention:} Different model architectures focus on different aspects of the input. Extraneous information that distracts one model may be ignored by others with different inductive biases.
    
    \item \textbf{Ranking-based validation:} The Bradley-Terry ranking aggregates preferences across models, naturally downweighting responses that appear confused or internally inconsistent due to noise.
    
    \item \textbf{Emergent signal extraction:} Through collective intelligence, the swarm extracts the true signal (the actual question) from noise, similar to how ensemble methods in signal processing reject outliers.
\end{enumerate}

This evaluation demonstrates that swarm intelligence not only improves baseline accuracy but fundamentally enhances robustness—a critical property for trustless, decentralized AI systems where adversarial or poorly-formed inputs are inevitable.

\subsubsection{Byzantine Fault Tolerance}

We test system performance with varying fractions of Byzantine nodes providing random or adversarial rankings:

\begin{figure}[h]
\centering
\begin{tikzpicture}
    \begin{axis}[
        width=12cm,
        height=7cm,
        xlabel
={Byzantine Node Fraction (\%)},
        ylabel={Accuracy (\%)},
        legend pos=north east,
        grid=major,
        xmin=0, xmax=50,
        ymin=20, ymax=85
    ]
    
    \addplot[color=blue, thick, mark=*] coordinates {
        (0,81.76) (10,79.8) (20,76.4) (30,71.2) (40,58.3) (50,42.1)
    };
    \addlegendentry{Fortytwo (with reputation)}
    
    \addplot[color=cyan, thick, mark=o] coordinates {
        (0,81.76) (10,75.2) (20,65.8) (30,52.3) (40,38.7) (50,28.4)
    };
    \addlegendentry{Fortytwo (no reputation)}
    
    \addplot[color=red, thick, mark=square] coordinates {
        (0,68.69) (10,52.3) (20,41.7) (30,35.2) (40,28.4) (50,24.1)
    };
    \addlegendentry{Majority Voting}
    
    \addplot[color=black, dashed] coordinates {(33.33,0) (33.33,85)};
    \addlegendentry{Theoretical Limit}
    
    \end{axis}
\end{tikzpicture}
\caption{Byzantine fault tolerance showing graceful degradation with malicious nodes}
\label{fig:byzantine_tolerance}
\end{figure}
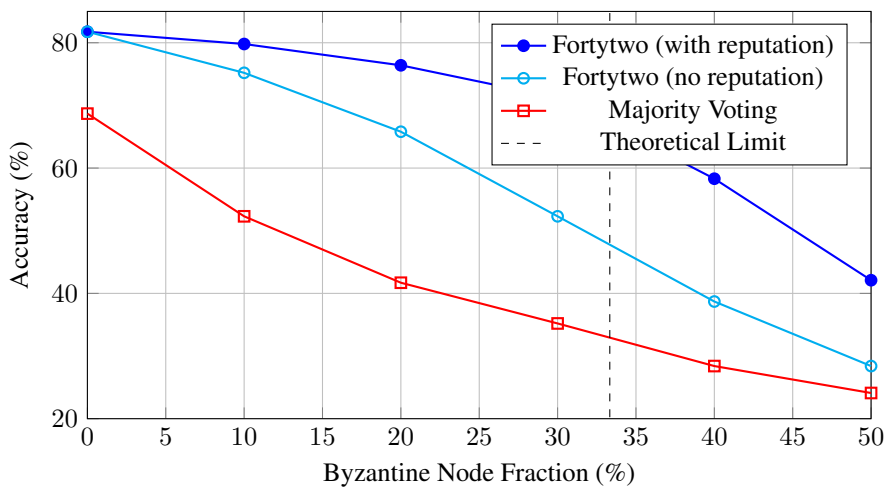

Key findings on Byzantine resilience:

\begin{itemize}
    \item \textbf{Exceeds theoretical limits:} Reputation weighting enables tolerance beyond the classical 1/3 Byzantine bound
    
    \item \textbf{Graceful degradation:} Performance decreases gradually rather than catastrophically failing
    
    \item \textbf{Reputation is crucial:} Reputation weighting provides significant additional resilience
    
    \item \textbf{Practical tolerance:} System maintains usable performance (>70\%) with up to 30\% Byzantine nodes
\end{itemize}

\subsection{Economic Analysis}

\subsubsection{Computational Efficiency}

We measure the computational overhead of our approach compared to alternatives. For the measurement of ranking consensus, we use our ranking sampling approach and the assumption that average ranking is 5\% of size compared to the response output tokens.

\begin{table}[h]
\centering
\caption{Computational cost analysis (relative to single model inference)}
\label{tab:computational_cost}
\begin{tabular}{lccc}
\toprule
\textbf{Method} & \textbf{Inference} & \textbf{Consensus} & \textbf{Total} \\
\midrule
Single Model & 1.0× & -- & 1.0× \\
Fortytwo (35 nodes) & 35.0× & 5.05× & 40.05× \\
ZKML & 1.0× & 10,000× & 10,001× \\
OPML & 1.0× & 100× & 101× \\
\bottomrule
\end{tabular}
\end{table}

Our approach achieves reasonable overhead compared to alternatives, with consensus formation requiring only 2-5× the cost of a single inference while improving the answer accuracy to the level of frontier model.

\subsection{Qualitative Analysis}

\subsubsection{Ranking Rationale Analysis}

Examining the reasoning chains generated during the pairwise comparison reveals a sophisticated evaluation. The models go over different troubling aspects of the response candidates. 

\textit{Example ranking rationale:}
\begin{quote}
Solution 1 lacks:\\\
- Incorrectly identifies A4 as KI based on an assumption about A5's decomposition, despite strong evidence for Y=I and A2=IF3.\\\
- Incorrectly calculates the molecular weight of A4 based on the incorrect identification of A4.\\\
- Ignores the strong evidence that A4 is a fluoride of Y, given the context of other fluorides.\\\
\\\
Solution 2 lacks:\\\
- Does not fully resolve the "bright-red" color of A1 and its decomposition at 293K, acknowledging it as an inconsistency but proceeding with the most robust numerical evidence.\\\
\\\
Solution 2 is better because it correctly identifies element Y and A2 based on the precise fluorine percentage, and then uses the properties of A4 (colorless concentrated solution) to correctly identify it as IF5. Solution 1 makes an unsupported leap to identify A4 as KI, which contradicts the context of binary fluorine compounds of element Y.\\\
\\\
\textbf{2}
\end{quote}

This nuanced evaluation demonstrates the system's ability to consider multiple quality dimensions and make context-appropriate decisions.

\subsubsection{Response Quality Examples}

We provide illustrative examples that highlight the quality improvements achieved through swarm consensus. To maintain both clarity and task complexity, we include selected samples from the SimpleBench benchmark.\citep{philip2024simplebench}:

\begin{table}[h]
\centering
\caption{Example responses showing quality improvement through peer-review consensus}
\label{tab:response_examples}
\begin{tabular}{p{7cm}p{4cm}p{4cm}}
\toprule
\textbf{Query} & \textbf{Majority Vote Selection} & \textbf{Fortytwo Selection} \\
\midrule
"A luxury sports-car is traveling north at 30km/h over a roadbridge, 250m long, which runs over a river that is flowing at 5km/h eastward. The wind is blowing at 1km/h westward... 1 hour later, the water-proof glove is (relative to the center of the bridge) approximately:
\newline\newline
A. 4km eastward \newline
B. $<$1 km northward \newline
C. $>$30km away north-westerly \newline
D. 30 km northward \newline
E. $>$30 km away north-easterly. \newline
F. 5 km+ eastward" & 
\textbf{Answer: F – about 5km eastward} of the bridge's centre. & 
\textbf{Answer: B. $<$1km northward} – in fact, essentially zero; the glove stays right where it fell, at the bridge's centre. \\
\midrule
"John is 24 and a kind, thoughtful and apologetic person. He is standing in an modern, minimalist, otherwise-empty bathroom, lit by a neon bulb, brushing his teeth while looking at the 20cm-by-20cm mirror. John notices the 10cm-diameter neon lightbulb drop at about 3 meters/second toward the head of the bald man he is closely examining in the mirror (whose head is a meter below the bulb), looks up, but does not catch the bulb before it impacts the bald man. The bald man curses, yells 'what an idiot!' and leaves the bathroom. Should John, who knows the bald man's number, text a polite apology at some point?" & 
\textbf{Answer: Yes. }\newline
If John feels responsible for the mishap, sending a brief, sincere apology is the courteous thing to do. A short message like “Hey, I’m really sorry about what happened in the bathroom. I hope you’re okay. I’ll be more careful next time.” acknowledges his part in the incident and gives the bald man a chance to let go of the frustration. The key point is that a polite text shows good character and makes it easier to clear the air. & 
\textbf{Answer: No – John shouldn’t text.}\newline
\newline
The “bald man” that John sees in the mirror is \textit{himself}.\newline
The neon bulb falls on the head that’s a meter below the bulb, which is John’s own head.\newline
When the bulb hits it, the “bald man” yells “what an idiot!” – that’s John shouting at himself.\newline
\newline
So there is no other person to apologize to; John is the bald man. \\
\bottomrule
\end{tabular}
\end{table}

As we see, a majority vote selects the invalid answer based on the model's pretraining bias. In contrast, swarm's peer ranked consensus picks the most prominent one, extracting a deeper valid signal.    

\section{Discussion}

\subsection{Why Pairwise Peer-Ranking Succeeds}

The exceptional performance of our pairwise ranking approach emerges from the convergence of multiple factors that synergistically amplify each other. Understanding these success factors provides information for the future development of collective intelligence systems.

\subsubsection{Cognitive Foundations}

The superiority of comparative over absolute judgment has deep roots in human cognition and extends naturally to artificial intelligence systems. When nodes evaluate two responses simultaneously, they engage in direct feature comparison rather than abstract quality assessment. This process activates what we term "contrastive attention"—the automatic highlighting of differences that might be overlooked in isolation.

Consider how an node evaluates factual accuracy: in absolute scoring, the node must maintain an internal standard of "accuracy" and measure each response against this abstract benchmark. In pairwise comparison, the node directly identifies the factual claims in each response and checks for contradictions or errors. This concrete and differential process is both more reliable and more interpretable.

The multitoken reasoning requirement amplifies these cognitive advantages by forcing systematic evaluation. The mean reasoning chain length of 127 tokens in our experiments represents substantial deliberation, equivalent to a thoughtful paragraph of analysis. This deliberation engages what dual-process theory calls System 2 thinking: slow, analytical, and logical, in contrast to the fast, intuitive System 1 that dominates single-token predictions \citep{kahneman2011thinking}. Having the reasoning chain length strictly limited greatly reduces hallucinations and speculation in the ranking process.

\subsubsection{Statistical Power of Tournament Aggregation}

The Bradley-Terry model's effectiveness stems from its optimal use of comparison information. Each pairwise comparison provides one bit of information (accounting for probabilistic outcomes), and the model extracts maximum likelihood estimates that best explain all observed comparisons. With $N$ items and $O(N^2)$ comparisons, the model can recover quality scores with precision proportional to $1/\sqrt{N}$, achieving statistical efficiency close to the Cramér-Rao lower bound.

The redundancy in our system, where multiple nodes evaluate overlapping comparison pairs, provides robustness through statistical averaging. Even if individual comparisons are noisy (which they inevitably are), the aggregate contains strong signal about relative quality. The law of large numbers ensures that random errors cancel out while systematic quality differences are preserved and amplified.

Furthermore, the Bradley-Terry model gracefully handles intransitive preferences that would break simple voting systems. If Response A beats B, B beats C, but C beats A (a preference cycle, which is quite observable in practice with LLMs), the model finds quality scores that best approximate these relationships in a least-squares sense, maintaining global consistency despite local inconsistencies.

\subsubsection{Emergence Through Diversity}

The heterogeneity of our node swarm creates complementary expertise that exceeds any individual's capabilities. Different models exhibit distinct strengths, and specialized models excel at domain-specific tasks. When these diverse perspectives are intelligently aggregated, the collective covers the weaknesses of each other.

This diversity also provides natural defense against adversarial attacks and biases. An adversarial example crafted to fool one model architecture often fails against others with different training data or inductive biases. Similarly, biases present in one model (e.g., verbosity preference) are counterbalanced by others with different biases. The ranking mechanism naturally identifies and upweights the most reliable evaluations for each specific query type.

The sampler variations within our swarms like temperature or nucleus sampling add another layer of diversity. Zero-temperature instances provide consistent, high-confidence evaluations while higher-temperature instances explore alternative perspectives. This exploration-exploitation balance ensures both reliability and creativity in the final consensus.

\subsubsection{Economic Incentive Alignment}

The reputation system creates an incentive structure that encourages honest and effortful participation. Unlike simple voting where all participants have equal weight regardless of quality, our system rewards competence with influence and earnings. This meritocracy naturally evolves toward higher quality as successful nodes gain resources to improve (through better hardware or tuning) while poor performers are marginalized.

The compute stake mechanism ensures that all participants have genuine capability, filtering out low-effort participants who might otherwise degrade system quality. The ongoing cost of maintaining multiple identities makes Sybil attacks economically unviable, while the reputation system's memory ensures that past performance influences future opportunities.

Importantly, the system aligns individual and collective interests. Nodes maximize their own rewards by providing accurate rankings that help the swarm identify the best responses. There's no incentive for strategic voting or manipulation because reputation updates are based on alignment with the final consensus, which emerges from the collective evaluation.

\subsection{Limitations and Challenges}

Despite our strong results, several limitations warrant careful consideration for practical deployment and future research.

\subsubsection{Latency Considerations}

The multiphase consensus process introduces latency that may be unacceptable for real-time applications. Our measurements show added end-to-end latency of at least 2-5 seconds for simple queries. This overhead comes from response generation parallelization limits, comparison evaluation time, and consensus formation computation.

For latency-sensitive applications, we envision a hybrid approach where initial response from the participating node with the highest reputation is returned immediately while consensus formation continues in the background. Users could receive progressive quality improvements as more nodes contribute evaluations. 

\subsubsection{Interpretability and Debugging}

While reasoning chains provide some interpretability, understanding why the swarm reached a particular consensus remains challenging. The Bradley-Terry optimization produces quality scores, but these are relative rankings without absolute meaning. When the system fails, identifying whether the problem lies in response generation, comparison evaluation, or aggregation requires a sophisticated analysis.

Future work should develop better visualization and debugging tools for swarm behavior. This could include attention analysis showing which response features influenced rankings, contribution tracking identifying which nodes most influenced consensus, and counterfactual analysis exploring how different node configurations would change outcomes.

In such a way, more grounded and unbiased ranker models can be tuned and trained, further improving quality and accuracy inside the swarm.

\subsection{Future Research Directions}

Our work opens numerous avenues for future investigation that could further advance decentralized AI systems.

\subsubsection{Theoretical Foundations}

While we demonstrate empirical success, deeper theoretical understanding would strengthen the approach. Key questions include:

\begin{itemize}
    \item What is the distribution of ranking accuracy? How effective is it over all possible input domains?
    \item How does node diversity quantitatively contribute to system performance?
    \item Can we prove robustness guarantees against specific attack classes?
    \item What are the optimal swarm sizes for different problem classes?
\end{itemize}

The relationship between our approach and other ensemble methods deserves exploration to identify fundamental principles of collective intelligence.

\subsubsection{Alternative Consensus Mechanisms}

While Bradley-Terry aggregation works well, other preference learning models might offer advantages. The Plackett-Luce model generalizes to rankings over multiple items simultaneously. Thurstone models assume Gaussian rather than logistic noise. Recent work on learning from human feedback suggests that more sophisticated preference models could better capture the nuances of quality assessment.

As for the protocol side, integration of cryptographic commitments scheme could ensure that nodes cannot change their evaluations after seeing others' rankings and/or responses, thus decreasing effective latency.

\subsubsection{Cross-Modal Extension}

Our framework currently focuses on text, but the principles extend naturally to other modalities. Vision-language models could evaluate image generation quality through pairwise comparison. Audio models could rank speech synthesis or music generation. Multimodal nodes could evaluate complex content combining text, images, and audio.

While such support requires the development of appropriate comparison templates and quality metrics for different modalities, the challenge lies in supporting the infrastructure for the orders of magnitude larger data transfer during the inference time.

\subsubsection{Human-AI Collaboration}

While our system operates autonomously, integration with human feedback could enhance performance for subjective or value-laden tasks or as a ground truth system. Humans could provide high-quality seed rankings that calibrate the swarm's evaluations. Expert humans could serve as high-reputation nodes for specialized domains. Conversely, the swarm could assist human decision-making by providing ranked options with explanations.

This human-in-the-loop approach raises questions about interface design, trust calibration, and appropriate task division. These questions become critical as AI systems increasingly support high-stakes human decisions.

\section{Conclusion}

Fortytwo represents a new approach in decentralized artificial intelligence, demonstrating that collective intelligence can exceed individual capabilities while maintaining practical deployability. Through the novel combination of distributed pairwise ranking, multitoken reasoning chains, compute stake mechanisms, and reputation-weighted consensus, we achieve performance that surpasses both traditional voting methods and challenges state-of-the-art monolithic models.

Our experimental results validate the theoretical promise of swarm intelligence for AI systems. The 17.21\% improvement over majority voting on GPQA Diamond and the remarkable robustness to free-form prompts demonstrate that properly orchestrated node swarms can achieve emergent intelligence beyond their individual components. The system's ability to maintain only 0.12\% performance degradation under adversarial conditions, compared to 6.20\% for individual models, suggests a new paradigm for building robust AI systems through diversity and validation rather than hardening individual models.

The compute stake mechanism addresses a fundamental challenge in decentralized systems, preventing Sybil attacks without creating plutocracy. By requiring proof of capability rather than proof of resources, we enable democratic participation while maintaining quality and security. The reputation system creates evolutionary pressure toward excellence, and the swarm naturally improves over time as successful nodes gain influence while poor performers are marginalized.

The broader implications of our work extend beyond the technical achievements. By enabling high-quality AI inference on heterogeneous hardware through collective intelligence, we contribute to the democratization of AI technology. Small organizations and individuals can participate in and benefit from advanced AI capabilities without massive infrastructure investments. The trustless nature of our consensus mechanism, combined with cryptographic guarantees and public auditability, addresses growing concerns about AI transparency and accountability.

Looking ahead, the principles demonstrated here, collective intelligence through diversity, quality through competition, and security through consensus, provide a template for future AI systems. As models continue to grow in size and capability, swarm-based approaches offer a path toward a sustainable and accessible AI infrastructure. The ability to achieve superhuman performance through collective evaluation suggests that the future of AI may lie not in ever larger individual models, but in orchestration of diverse, specialized nodes.

The Fortytwo Protocol is more than a technical solution; it represents a vision for AI development that is open, robust, and aligned with human values. By demonstrating that decentralized systems need not sacrifice quality for democratization, we hope to inspire further research into collective AI intelligence. The convergence of swarm intelligence, blockchain technology, and advanced language models opens possibilities that we are only beginning to explore.

As we approach the threshold of artificial general intelligence, questions of control, access, and accountability become paramount. Our work suggests that these challenges might be best addressed not through centralized control but through decentralized cooperation, not through individual superhuman AI but through collective intelligence that remains comprehensible and controllable. The swarm, with its redundancy, diversity, and consensus building, may be more aligned with human values than any individual artificial intelligence, no matter how advanced.

In the grand challenge of creating beneficial artificial intelligence, we offer evidence that the whole can indeed be greater than the sum of its parts and that this collective intelligence can be achieved today with existing technology, waiting only for adoption and refinement by the global community of researchers, developers, and users who will shape the future of AI.

\bibliographystyle{plainnat} 
\bibliography{references} 

\end{document}